\documentclass[twocolumn,10pt]{article}

\usepackage[margin=1in]{geometry}
\usepackage[T1]{fontenc}
\usepackage[utf8]{inputenc}

\usepackage{amsmath,amssymb}

\usepackage{graphicx}
\usepackage{booktabs}
\usepackage{multirow}
\usepackage{float}
\usepackage{subcaption}   % use this; DO NOT use subfigure

\usepackage{multicol}

\usepackage[table,xcdraw]{xcolor}

\usepackage{arydshln}

\usepackage{listings}

\usepackage[hidelinks]{hyperref}

\usepackage{float} 
\usepackage{tcolorbox}
\usepackage{multirow}
\usepackage{graphicx}
\usepackage{booktabs}
\usepackage{xcolor}
\usepackage{colortbl}
\usepackage{mdframed}
\usepackage{listings}
\usepackage{xcolor}

\lstdefinelanguage{Prompt}{
  morekeywords={Output, JSON, label, return},
  sensitive=false,
  morecomment=[l]{//},
  morestring=[b]",
}

\title{MammoWise: Multi-Model Local RAG Pipeline for Mammography Report Generation}

\author{
  Raiyan Jahangir, Nafiz Imtiaz Khan, Amritanand Sudheerkumar, Vladimir Filkov \\
  \textit{University of California, Davis, CA, USA}
}
\date{} % keep empty for arXiv

\begin{document}
\maketitle

\begin{abstract}
Screening mammography is high-volume, time-sensitive, and documentation-heavy. Radiologists must translate subtle visual findings into consistent BI-RADS assessments, breast-density categories, and narrative reports. Although recent Vision Language Models (VLMs) make image-to-text reporting plausible, many demonstrations rely on closed, cloud-hosted systems (raising privacy, cost, and reproducibility concerns) or on tightly coupled, task-specific architectures that are hard to adapt across models, datasets, and workflows. We present \textit{MammoWise}, a local, multi-model pipeline that turns open-source VLMs into clinically-styled mammogram report generators and multi-task classifiers. \textit{MammoWise} accepts any Ollama-hosted VLM and any mammography dataset, supports zero-shot, few-shot, and Chain-of-Thought prompting, and includes an optional multimodal Retrieval Augmented Generation (RAG) mode that retrieves image-text exemplars from a vector database to provide case-specific context. To illustrate the utility of \textit{MammoWise} in complex use cases, we evaluate three open VLMs (\textit{MedGemma}, \textit{LLaVA-Med}, \textit{Qwen2.5-VL}) on VinDr-Mammo and DMID datasets, and assess report quality (BERTScore, ROUGE-L), BI-RADS classification (5-class), breast density (ACR A–D), and key findings (mass, calcification, asymmetry). Across models and datasets, report generation is consistently strong. It generally improves with few-shot prompting and RAG, whereas prompting-only classification is feasible but highly sensitive to the (dataset, model, prompt) combination. To improve reliability, we added parameter-efficient fine-tuning (QLoRA) to \textit{MedGemma} on VinDr-Mammo, yielding substantial gains, including BI-RADS accuracy of 0.7545, density accuracy of 0.8840, and calcification accuracy of 0.9341, while preserving report quality. Overall, \textit{MammoWise} provides a practical, extensible scaffold for deploying and studying local VLMs for mammography reporting, spanning prompting, RAG, and fine-tuning within a unified, reproducible workflow.
\end{abstract}

\section{Introduction}
Screening mammography is among the most widely used imaging modalities in modern preventive care \cite{lashof2001mammography}. Every study requires careful inspection of bilateral views, such as Cranio-Caudal (CC) and Medio-Lateral Oblique (MLO), identification of subtle findings (masses, asymmetries, calcifications), and conversion of visual evidence into standardized clinical language: BI-RADS evaluation and recommendation, breast-density assessment, and a structured narrative report \cite{jahangir2025mammo}. This translation step is essential, as reports drive follow-up imaging, biopsies, and longitudinal follow-up \cite{mabotuwana2019automated}. Still, it is also labor-intensive and vulnerable to variability in phrasing, completeness, and coding-relevant details. As imaging volumes continue to grow, tools that can help draft consistent, clinically styled mammogram reports have immediate practical value, without compromising privacy or introducing unsafe hallucinations \cite{ji2023survey}.

VLMs offer a promising direction because they can jointly reason over images and text and generate natural-language output \cite{zhang2024vision}. In principle, a VLM could ingest mammogram views and produce (i) a narrative report that follows radiology conventions and (ii) structured labels such as BI-RADS and density. However, in practice, two friction points limit the usefulness of the real world. First, many of the strongest demonstrations of report generation use large, closed, cloud-hosted models \cite{raiaan2024review}. For clinical imaging, this creates deployment barriers. For example, protected health information may be transmitted off-premises, operational costs can be high, and results are difficult to reproduce or audit. Second, open-source VLMs can be run locally, but "out of the box,” they are often not specialized for mammography and may produce clinically implausible details unless carefully guided \cite{jia2022visual}. Bridging this gap typically requires either prompt engineering \cite{giray2023prompt} (lightweight but unreliable for fine-grained classification) or fine-tuning \cite{jahangir2023comparative, majib2021vgg} (more reliable but model- and dataset-dependent and often expensive).

Existing multimodal mammography-focused systems illustrate these trade-offs. Several lines of work emphasize specialized architectures or training recipes optimized for specific tasks (e.g., malignancy, breast density, or lesion detection), while others focus on reporting using proprietary models \cite{sexauer2023diagnostic}. As a result, practitioners and researchers who want to (a) run locally, (b) compare multiple open VLMs, and (c) evaluate multiple adaptation strategies (prompting, retrieval-augmented prompting, and fine-tuning) often face fragmented codebases, inconsistent experimental protocols, and limited support for producing complete, clinically styled reports rather than only labels.

This paper takes a different stance. Instead of proposing yet another bespoke model, we introduce \textit{MammoWise}, a multi-model, locally runnable pipeline that makes open VLMs usable for mammogram report generation and associated classification tasks. Conceptually, MammoWise is a modular one-stop shop, comparable to middleware/IDE, that provides a unified interface to launch local VLMs, configure multimodal RAG, and run standardized evaluations across multiple mammography use cases and metrics. The design goal is to enable practical, reproducible experimentation under realistic constraints. The pipeline can plug in any Ollama-hosted VLM \cite{marcondes2025using}, ingest standard mammography datasets, and run a consistent end-to-end workflow that includes preprocessing, prompt templating, output parsing, and evaluation. \textit{MammoWise} supports zero-shot \cite{pourpanah2022review}, few-shot \cite{wang2020generalizing}, and Chain-of-Thought (CoT) prompting \cite{wei2022chain}. It also adds an optional RAG \cite{patidar2024transparency} mode that selects image-text exemplars from a multimodal vector database to provide a customized context to the input mammogram. When prompting alone is insufficient, especially for stable, high-fidelity classification, we also support Parameter-Efficient Fine-Tuning (PEFT) \cite{fu2023effectiveness} (QLoRA), which adapts an open VLM to the target label space without full model retraining.

We evaluate \textit{MammoWise} across two datasets (VinDr-Mammo \cite{nguyen2023vindr} and DMID \cite{oza2024digital}) and three open VLMs (\textit{MedGemma} \cite{sellergren2025medgemma}, \textit{LLaVA-Med} \cite{li2023llava}, Qwen2.5-VL \cite{bai2025qwen2}). The results highlight a consistent pattern. Report generation is robust and improves further with few-shot prompting and RAG. At the same time, prompting-only classification is highly variable across model/dataset/prompt choices. To reduce this sensitivity, we fine-tune \textit{MedGemma} on VinDr-Mammo using QLoRA \cite{dettmers2023qlora} and observe substantial improvements in classification quality, reaching 0.7545 BI-RADS accuracy, 0.8840 density accuracy, and 0.9341 calcification accuracy (F1-score 0.9313) while maintaining strong report-generation behavior. These findings suggest a pragmatic division of labor: prompting and retrieval are often sufficient to draft clinically styled reports, but dependable structured classification benefits from lightweight fine-tuning.

\subsection*{Contributions}
This paper makes the following contributions:
\begin{itemize}
    \item A local, multi-model pipeline (\textit{MammoWise}) that turns open VLMs into mammogram report generators and multi-task classifiers across datasets and prompting regimes.
    \item A multimodal RAG workflow for case-specific few-shot context using an image-text vector database.
    \item A systematic comparison of prompting, RAG, and QLoRA fine-tuning across models and datasets, with unified evaluation for both narrative report quality and clinical labels.
    \item Evidence that parameter-efficient fine-tuning can substantially stabilize and improve classification performance in this setting.

\end{itemize}
Based on these contributions, we formulate the following research questions:

\vspace{0.15cm}
\begin{mdframed}[backgroundcolor=blue!5, linecolor=black!70, roundcorner=10pt]
\textbf{RQ}$_1:$ How well do local medical VLMs generate structured reports under prompting vs RAG? \newline
\textbf{RQ}$_2$: When does lightweight fine-tuning outperform prompting/RAG for classification? \newline
\end{mdframed}

The remainder of the paper reviews related work, describes the \textit{MammoWise} architecture and experimental design, presents results and analyses across adaptation strategies, and closes with practical implications and limitations.

\begin{figure*}[!ht]
    \centering
    \begin{minipage}{\textwidth}
        \centering
        \includegraphics[width=\linewidth,height=100mm]{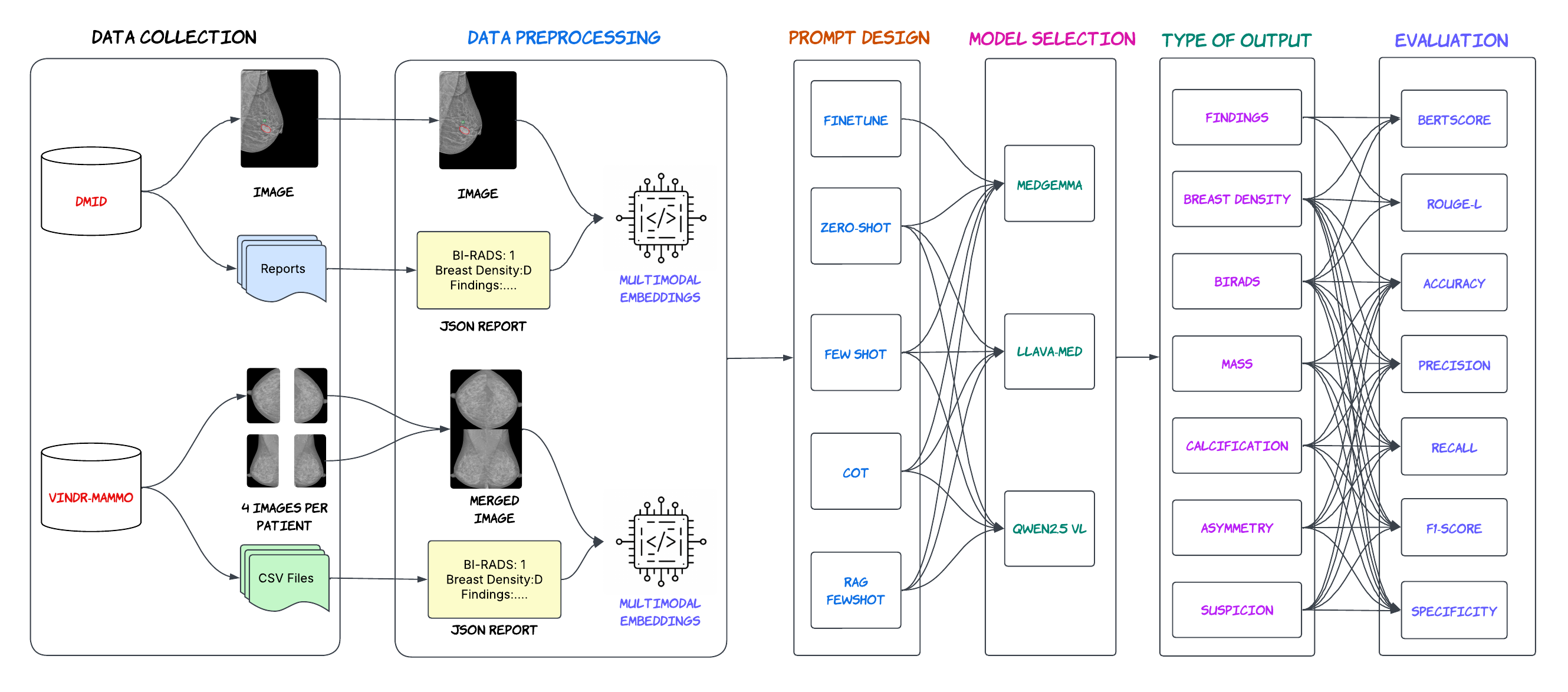}
        \caption{Overall methodology of the study}
        \label{methodology}
    \end{minipage}
\end{figure*}

\section{Literature Background}
Several recent studies have explored how VLMs can assist radiologists with mammogram interpretation. Haver et al. \cite{haver2024use} investigated the potential of \textit{ChatGPT} to determine BI-RADS assessments from mammogram images. Pesapane et al. \cite{pesapane2025preliminary} used \textit{GPT-4} to generate mammogram reports and identify abnormalities, such as masses and microcalcifications. While these works demonstrated that large, closed-source models can describe mammogram findings, they also reported hallucinations, low sensitivity and specificity, and the inherent privacy and cost issues associated with cloud-based models.

Other work has investigated CLIP-style models as backbones for mammography. Moura et al. \cite{de2024unlocking} evaluated several Contrastive Language-Image Pretraining (CLIP) variants-\textit{CLIP} \cite{li2025closer}, \textit{BiomedCLIP} \cite{zhang2023biomedclip}, and \textit{PubMedCLIP} \cite{eslami2023pubmedclip} for breast-density and BI-RADS classification. Khan et al. \cite{urooj2024knowledge} used \textit{MedCLIP} \cite{wang2022medclip} to construct diverse teaching cases for radiology education via image-text retrieval, exploring zero-shot, few-shot, and supervised prompting scenarios.

Several specialized VLM architectures have also been proposed. Chen et al. introduced \textit{LLaVA-Mammo} \cite{chen2025llava}, a fine-tuned variant of \textit{LLaVA} \cite{liu2023visual} tailored for BI-RADS, breast density, and malignancy classification. Increasing the language model size from 7B to 13B parameters improved density accuracy from 72.1 to 76.6 and malignancy AUC from 0.687 to 0.723. Their later \textit{LLaVA-MultiMammo} model \cite{chen2025llava2} extended capabilities to breast cancer risk prediction while retaining core classification tasks. Ghosh et al. \cite{ghosh2024mammo} proposed \textit{MammoCLIP}, a VLM combining CNN-based vision \cite{ramadan2023enhancing} and \textit{BioBERT} \cite{lee2020biobert} language encoders trained on mammogram-report pairs. Their model supports the classification of malignancy, density, mass, and calcification, and provides visual explanations via a novel feature attribution method (Mammo-Factor). Jain et al. \cite{jain2024mmbcd} developed \textit{MMBCD}, a multimodal model that integrates mammogram images and clinical histories using ViT and \textit{RoBERTa}-based embeddings \cite{liu2019roberta}. Cao et al. \cite{cao2025mammovlm} introduced \textit{MammoVLM}, a \textit{GLM-4-9B}-based VLM with a dedicated visual encoder, fine-tuned to provide diagnostic assistance at a level comparable to a junior radiologist.

Very few works provide a single reusable pipeline that supports both report generation and multi-task labels across models/datasets. Our work addresses this gap by framing \textit{MammoWise} as a reusable pipeline rather than a single fixed model, and by evaluating its behavior across multiple VLMs, datasets, and adaptation strategies.

\section{Methodology}
The study's overall methodology is shown in Figure \ref{methodology} and discussed in detail in the following subsections.

\subsection{Data Collection}
We used two mammography datasets for our work. The primary dataset, VinDr-Mammo, was prepared by Nguyen et al. \cite{nguyen2023vindr} and is used for most experiments, including fine-tuning. The second dataset is the DMID dataset prepared by Oza et al. \cite{oza2024digital}.

The VinDr-Mammo dataset consists of 20,000 images from 5,000 Vietnamese patients, with four images per patient (two per breast): Cranio-Caudal (CC) and Medio-Lateral Oblique (MLO) for both left and right breasts. The dataset also includes metadata describing acquisition devices, study numbers, and image identifiers, as well as breast-level information like tissue density (ACR A/B/C/D) and BI-RADS score (1–5), and abnormal findings such as masses, calcifications, and other abnormalities. The dataset is publicly available via \href{https://physionet.org/content/vindr-mammo/1.0.0/}{PhysioNet \newline(https://physionet.org/content/vindr-mammo/1.0.0/)}.

DMID is a relatively small but lesion-enriched dataset that contains 510 mammogram images from India, of which 300 include abnormalities. Each image is paired with a corresponding text report describing breast density, BI-RADS score, and any abnormal findings. This image-report pairing makes DMID particularly suitable for evaluating report-generation behavior and prompting strategies. The dataset is accessible via \href{https://figshare.com/authors/Parita_Oza/17353984}{Figshare \newline(https://figshare.com/authors/Parita\_Oza/17353984)}.

\subsection{Data Preprocessing}
To prevent memory overload and ensure compatibility with the VLMs, we resize all images from both datasets to a fixed size of $512 \times 512$ pixels. For VinDr-Mammo, we then merge the four views (right/left CC and right/left MLO) of each patient into a single composite image, as shown in Figure \ref{methodology}, arranged symmetrically: CC views are side by side at the top, and MLO views are side by side at the bottom. This layout mimics the way radiologists routinely inspect mammograms, comparing the right and left breasts on both projections to assess asymmetries and subtle patterns \cite{sasikala2019breast,chotai2020breast}.

From the 20,000 single-view images, this merging process yields 5,000 composite images, one per patient. For each merged image, we generate a corresponding report, yielding 5,000 composite image-report pairs. In the report:
\begin{itemize}
    \item Breast density is taken directly from the breast-level labels.
    \item For BI-RADS, when different views have different categories, we take the highest BI-RADS across the four views as the patient-level label, reflecting clinical caution.
    \item The findings of all views are combined into a single sentence summarizing the presence and location of masses, calcifications, asymmetries, and other abnormalities.
    \item We define a field of derived \emph{suspicion} that indicates whether the case is labeled ``healthy'' (BI-RADS 1), ``benign'' (BI-RADS 2--3), or ``suspicious'' (BI-RADS 4-5). Importantly, this label reflects \emph{radiologic suspicion} implied by BI-RADS, not biopsy-confirmed cancer.
\end{itemize}

For the DMID dataset, there were already 510 corresponding reports for each image. We extract key information, such as ``breast density'', ``BI-RADS'', and ``findings'', from those reports and store them in JavaScript Object Notation (JSON) format.

For RAG implementation, we convert these image-report pairs into multimodal embeddings. We divide both the data into an 80:20 train-test split. For VinDr-Mammo, it is a patient-level split, whereas in DMID, it is an image-level split. The image and text embeddings are generated using a multimodal OpenCLIPEmbedding function, which applies a Vision Transformer model \cite{jahangir2023performance} for image encoding and a text encoder for report encoding. The resulting embeddings are stored as semantic indexes in ChromaDB \cite{lavanya2024advanced}, a vector database used for RAG retrieval in our experiments.

\begin{figure*}[!ht]
    \centering
    \begin{minipage}{\textwidth}
        \centering
        \includegraphics[width=\linewidth,height=80mm]{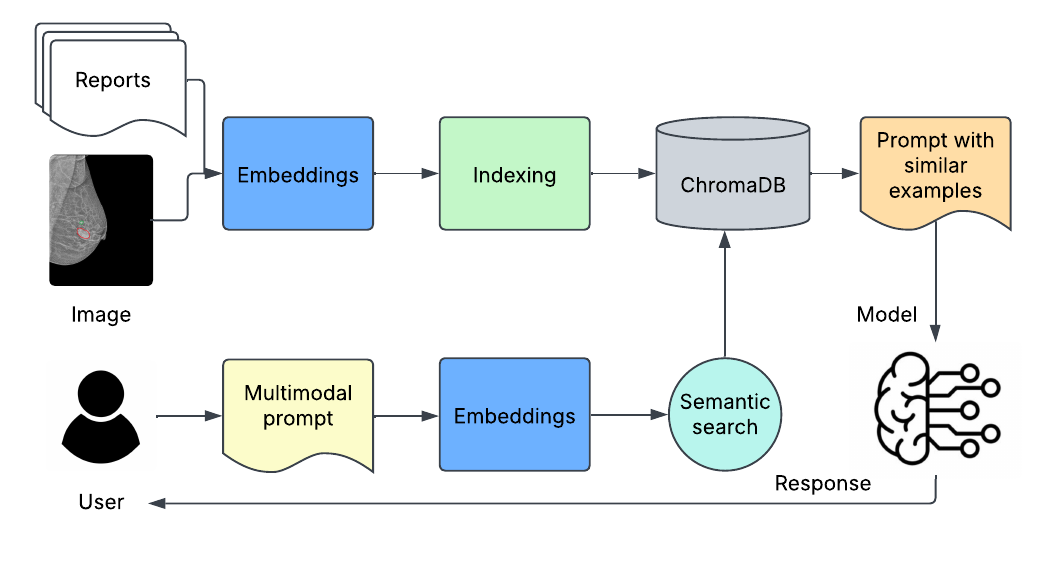}
        \caption{RAG pipeline used in this study}
        \label{rag}
    \end{minipage}
\end{figure*}

\subsection{Model Selection}
We selected three open-source VLMs available in Ollama \cite{marcondes2025using}, a framework that supports local hosting of multimodal language models. Each of these models can take an image and a text prompt as input and generate text as output. Because we work with medical data, we prioritize models that have been instruction-tuned on medical or biomedical content when possible. The three models are \textit{MedGemma}, \textit{LLaVA-Med}, and \textit{Qwen2.5-VL}, described below.

\textbf{MedGemma:} \textit{MedGemma} is a VLM created by Google \cite{sellergren2025medgemma}, derived from the \textit{Gemma 3} family \cite{team2025gemma}. It is specifically trained on medical data from radiology, pathology, and dermatology to generate medically grounded text based on multimodal input. It comes in 4B and 27B parameter variants; we use the 4B model to keep hardware requirements reasonable.

\textbf{LLaVA-Med:} \textit{LLaVA-Med} \cite{li2023llava} is a fine-tuned version of the general-purpose \textit{LLaVA} model \cite{liu2023visual}, trained on biomedical datasets. The base \textit{LLaVA} architecture combines a vision encoder with a Vicuna-based language backbone \cite{touvron2023llama}. We use the 7B-parameter \textit{LLaVA-Med} variant.

\textbf{Qwen2.5-VL:} \textit{Qwen2.5-VL} \cite{bai2025qwen2} is an improved version of the \textit{Qwen2} multimodal model \cite{wang2024qwen2}. It also has 7B parameters. Although it is not specifically tuned on medical data, it is trained to extract and reason about text and objects from images, making it a plausible candidate for detecting abnormalities in mammogram images.

\subsection{Prompt Design}
Prompt engineering is the process of creating customized instructions for LLMs and VLMs to guide them toward appropriate, well-structured responses \cite{zamfirescu2023johnny}. In our setting, the prompts include textual instructions and one or more mammogram images. To answer RQ1, we focus on three widely used prompting styles: zero-shot learning \cite{pourpanah2022review, khan2026intelligent}, few-shot learning \cite{wang2020generalizing}, and Chain-of-Thought (CoT) prompting \cite{wei2022chain}.

In the \textbf{zero-shot} setting, the model is given only high-level instructions about its role (``You are a board-certified breast radiologist…''), the image layout (4-view arrangement), and the desired output format (JSON with fields for breast density, findings, BI-RADS, and suspicion). For \textbf{few-shot} prompting, we augment the zero-shot template with five fixed examples of image-report pairs that illustrate the desired JSON structure and reporting style. These examples demonstrate how to map inputs to outputs. For \textbf{CoT prompting}, we modify the prompt so that the model emulates a radiologist's reasoning process step by step: first assess breast density, then identify findings, then determine whether the breast has suspicion of malignant tumors or not, and finally assign a BI-RADS category.

% Please add the following required packages to your document preamble:
% \usepackage{booktabs}

\begin{table*}[!ht]
\centering
\small
\caption{BI-RADS class counts before and after rebalancing for fine-tuning.}
\label{tab:BI-RADS_balance}
\begin{tabular}{crrl}
\toprule
\textbf{BI-RADS} & \textbf{Original} & \textbf{After} & \textbf{Action} \\
\midrule
1 & 3331 & 500  & Downsample (random) \\
2 & 1167 & 500  & Downsample (random) \\
3 & 242  & 500  & Augment (flip + scale) \\
4 & 205  & 500  & Augment (flip + scale) \\
5 & 55   & 200  & Augment (flip + scale + translation; capped) \\
\midrule
\textbf{Total} & \textbf{5000} & \textbf{2200} & \\
\bottomrule
\end{tabular}
\end{table*}

All prompt templates used in this study are provided in the Appendix.

\subsection{Experimental Design}
All experiments are conducted on a local machine, and each VLM is served via Ollama. We set the temperature parameter to 0 throughout, because mammogram reporting is a high-stakes task where we prioritize deterministic, low-variance outputs over creative variation \cite{renze2024effect}. Our experiments fall into three main configurations.

\subsubsection{Base Configuration}
In the base setting, we provide the model only with the prompt (zero-shot, few-shot, or CoT) and the images. The model's responses are captured and stored locally. This configuration isolates the effect of prompting alone, without additional retrieval or fine-tuning.

\subsubsection{RAG Configuration}
Hallucination-plausible but incorrect or fabricated content is a significant concern for VLMs \cite{rawte2023survey}. For mammogram reports, hallucinations are unacceptable. Explainability is another key requirement: clinicians want to understand \emph{why} a model produced a particular conclusion ~\cite{amann2020explainability,patidar2024transparency}. One practical way to improve grounding and interpretability is to use Retrieval-Augmented Generation (RAG) ~\cite{rag_2, rag_3, rag_1}, where the model is given relevant supporting examples or documents as context alongside the user query. A recent Nature article \cite{mesinovic2025explainability} also highlights RAG as a promising mechanism for improving transparency in medical generative models.

To enable the RAG, we design a pipeline that dynamically retrieves examples and feeds them into the prompt (Figure \ref{rag}). When a new image and prompt are provided, they are embedded in a multimodal space and sent to the ChromaDB vector database. The retrieval index is built only from training data with patient-level separation to avoid leakage. We then retrieve the five most similar image–report pairs using cosine similarity. These retrieved examples are inserted into the zero-shot prompt template, yielding a dynamic, context-aware few-shot prompt whose examples are semantically close to the current query. This is the \textbf{RAG} configuration.

\subsubsection{Fine-Tune Configuration}
Prompting alone leaves the model weights unchanged. The model does not gain new knowledge, which can limit performance. Fine-tuning can improve accuracy and robustness, but fine-tuning large VLMs to their full extent is often impractical on commodity hardware. Parameter-efficient fine-tuning \cite{fu2023effectiveness,sung2022vl} addresses this by training only a small set of adapter parameters while freezing the base model.

We fine-tune \textit{MedGemma}, the smallest of our candidate medical VLMs, on VinDr-Mammo using a class-rebalanced training subset constructed via data augmentation \cite{goceri2023medical}. The original cohort exhibits a strong BI-RADS imbalance (Table \ref{tab:BI-RADS_balance}). To mitigate this, we downsampled the majority classes (BI-RADS 1 and 2) to 500 images each by random sampling, and augmented the minority classes using simple geometric transforms commonly used in previous mammography work (flipping, scaling, and translation) \cite{blahova2025neural, oza2022image}. When applying horizontal flips, we update any laterality-sensitive report fields to preserve semantic consistency. We increased the number of BI-RADS 3 and 4 cases to 500 each. For BI-RADS 5, we increased the number of images from 55 to 200 (rather than 500) to respect a conservative augmentation cap reported in previous work (not exceeding 300\% augmentation) \cite{jimenez2024gan}. This yields a final fine-tuning set of 2,200 images. We loaded the pre-trained \textit{MedGemma} model from HuggingFace \cite{jain2022hugging} and applied Quantized Low-Rank Adaptation (QLoRA) \cite{dettmers2023qlora}, an efficient variant of LoRA \cite{hu2022lora}. QLoRA freezes the base weights and inserts low-rank adapter matrices into selected layers; only these adapters are trained, and at inference, their outputs are combined with the frozen weights to produce predictions. We fine-tune under two output formats: \textit{multi-task generation}, where the model outputs the full JSON report containing all target fields in a single response, and \textit{single-task generation}, where the model is prompted to output only one target field at a time. Unless stated otherwise, we report results for multiple checkpoints (multi-task: 3/6/10/15 epochs; single-task: 6/10 epochs) to capture task-dependent and potentially non-monotonic training effects. We trained with different numbers of epochs to observe how performance scales with them. The hyperparameters are as follows:
\newline
\begin{lstlisting}[]
Temperature: 0
Lora Alpha: 16
Lora Dropout: 0.05
Bias: None
Learning Rate: 2e-4
Batch size: 1
Optimizer: Adamw_bnb_8bit
Gradient Accumulation Steps: 8
Optimization: Qlora
\end{lstlisting}

\subsection{Evaluating Model Response}
We evaluate models along two main axes: classification performance and generation performance.

For \textbf{classification}, we consider three multi-class tasks (BI-RADS, breast density, and suspicion) and three binary tasks (presence or absence of mass, calcification, and asymmetry), reflecting abnormalities that radiologists routinely prioritize in mammogram interpretation. We report macro-averaged accuracy, precision, recall, F1-score, and specificity. Given class imbalances in both datasets, macro averaging ensures that minority classes receive equal weight \cite{jahangir2023brain, khan2020prediction}. This choice aligns with our study goal of comparing reliability across rare and common labels under prompting, RAG, and lightweight fine-tuning. However, macro-averaged improvements do not necessarily imply well-calibrated performance under the imbalanced distribution. Rather, calibration-oriented evaluation would be required for deployment-facing conclusions.

For \textbf{generation}, we assess whether the model's JSON output resembles human-written mammogram reports. We focus on BI-RADS descriptions, breast density descriptions, and findings. We compute BERTScore \cite{zhang2019bertscore} and ROUGE-L \cite{lin2004rouge}, since both capture semantic similarity rather than exact n-gram matches. BERTScore measures similarity in embedding space, while ROUGE-L evaluates the longest common subsequence between generated and reference text.

\section{Results \& Discussion}

% \input{tables/result_1}
% \input{tables/result_2}
% \input{tables/result_3}
% \input{tables/zeroshot}
% \input{tables/fewshot}
% \input{tables/COT}
% \input{tables/rag_fewshot}
% \input{tables/medgemma}
% \input{tables/comparison}

% \input{tables/rag_fewshot_vindr_class}
% \input{tables/rag_fewshot_dmid}
% Please add the following required packages to your document preamble:
% \usepackage{multirow}
% \usepackage{graphicx}
% \usepackage{booktabs}
% \usepackage{xcolor}
% \usepackage{colortbl}

\definecolor{headerblue}{RGB}{220, 230, 242}
\definecolor{rowgray}{RGB}{245, 245, 245}

\begin{table*}[!ht]
\centering
\caption{Performance of the models on VinDr-Mammo dataset without and with RAG}
\label{new_vindr_with_rag}
\resizebox{\textwidth}{!}{%
\begin{tabular}{llccccc}
\toprule
\multirow{2}{*}{\textbf{Task}} &
  \multirow{2}{*}{\textbf{Evaluation Parameters}} &
  \multicolumn{3}{c}{\textbf{Without RAG}} &
  \multicolumn{2}{c}{\textbf{With RAG}} \\
\cmidrule(lr){3-5} \cmidrule(lr){6-7}
 & & \textbf{Best Result} & \textbf{Prompt} & \textbf{Model} & \textbf{Result} & \textbf{Model} \\
\midrule

\multirow{7}{*}{BI-RADS}
 & Accuracy    & 0.2909 & Fewshot  & LLAVA-Med & 0.319  & MedGemma  \\
 & Precision   & 0.4121 & Fewshot  & Qwen2.5VL & 0.4508 & Qwen2.5VL \\
 & Recall      & 0.2909 & Fewshot  & LLAVA-Med & 0.319  & MedGemma  \\
 & F1-score    & 0.2217 & Fewshot  & LLAVA-Med & 0.2756 & Qwen2.5VL \\
 & Specificity & 0.8165 & Fewshot  & LLAVA-Med & 0.8265 & MedGemma  \\
 & BERTScore   & 0.9313 & Fewshot  & Qwen2.5VL & 0.9401 & MedGemma  \\
 & ROUGE-L     & 0.3593 & Fewshot  & Qwen2.5VL & 0.4347 & MedGemma  \\
\midrule

\multirow{7}{*}{Breast Density}
 & Accuracy    & 0.7832 & Fewshot   & Qwen2.5VL & 0.8068 & Qwen2.5VL \\
 & Precision   & 0.7923 & Fewshot   & LLAVA-Med & 0.8185 & Qwen2.5VL \\
 & Recall      & 0.7832 & Fewshot   & Qwen2.5VL & 0.8068 & Qwen2.5VL \\
 & F1-score    & 0.69   & Fewshot   & MedGemma  & 0.7537 & Qwen2.5VL \\
 & Specificity & 0.7754 & Zeroshot  & MedGemma  & 0.8082 & LLAVA-Med \\
 & BERTScore   & 0.9788 & Fewshot   & Qwen2.5VL & 0.9808 & Qwen2.5VL \\
 & ROUGE-L     & 0.8643 & Fewshot   & Qwen2.5VL & 0.874  & Qwen2.5VL \\
\midrule

\multirow{2}{*}{Findings}
 & BERTScore   & 0.9053 & Fewshot  & Qwen2.5VL & 0.9103 & MedGemma  \\
 & ROUGE-L     & 0.409  & Fewshot  & LLAVA-Med & 0.4652 & Qwen2.5VL \\
\midrule

\multirow{5}{*}{Calcification}
 & Accuracy    & 0.8391 & Fewshot   & Qwen2.5VL & 0.8591 & Qwen2.5VL \\
 & Precision   & 0.865  & Fewshot   & MedGemma  & 0.8409 & Qwen2.5VL \\
 & Recall      & 0.8391 & Fewshot   & Qwen2.5VL & 0.8591 & Qwen2.5VL \\
 & F1-score    & 0.7679 & Fewshot   & Qwen2.5VL & 0.8432 & Qwen2.5VL \\
 & Specificity & 0.5477 & Zeroshot  & Qwen2.5VL & 0.6879 & MedGemma  \\
\midrule

\multirow{5}{*}{Mass}
 & Accuracy    & 0.6218 & CoT      & LLAVA-Med & 0.6582 & Qwen2.5VL \\
 & Precision   & 0.7647 & Fewshot  & Qwen2.5VL & 0.7041 & Qwen2.5VL \\
 & Recall      & 0.6218 & CoT      & LLAVA-Med & 0.6582 & Qwen2.5VL \\
 & F1-score    & 0.5016 & Zeroshot & MedGemma  & 0.5726 & Qwen2.5VL \\
 & Specificity & 0.5211 & Fewshot  & MedGemma  & 0.5845 & MedGemma  \\
\midrule

\multirow{5}{*}{Asymmetry}
 & Accuracy    & 0.8915 & Zeroshot & LLAVA-Med & 0.8664 & Qwen2.5VL \\
 & Precision   & 0.7611 & Fewshot  & LLAVA-Med & 0.848  & LLAVA-Med \\
 & Recall      & 0.8915 & Zeroshot & LLAVA-Med & 0.8664 & Qwen2.5VL \\
 & F1-score    & 0.7875 & Zeroshot & MedGemma  & 0.8373 & Qwen2.5VL \\
 & Specificity & 0.5215 & CoT      & LLAVA-Med & 0.5951 & Qwen2.5VL \\
\midrule

\multirow{5}{*}{Suspicion}
 & Accuracy    & 0.6841 & Fewshot  & Qwen2.5VL & 0.735  & Qwen2.5VL \\
 & Precision   & 0.7841 & Fewshot  & Qwen2.5VL & 0.7282 & Qwen2.5VL \\
 & Recall      & 0.6841 & Fewshot  & Qwen2.5VL & 0.735  & Qwen2.5VL \\
 & F1-score    & 0.5897 & Fewshot  & LLAVA-Med & 0.6961 & Qwen2.5VL \\
 & Specificity & 0.5347 & CoT      & MedGemma  & 0.617  & Qwen2.5VL \\
\bottomrule
\end{tabular}%
}
\end{table*}
% Please add the following required packages to your document preamble:
% \usepackage{multirow}
% \usepackage{graphicx}
% \usepackage{booktabs}

\begin{table*}[!ht]
\centering
\caption{Performance of the models on DMID dataset without and with RAG}
\label{new_dmid_with_rag}
\resizebox{\textwidth}{!}{%
\begin{tabular}{llccccc}
\toprule
\multirow{2}{*}{\textbf{Task}} &
  \multirow{2}{*}{\textbf{Evaluation Parameters}} &
  \multicolumn{3}{c}{\textbf{Without RAG}} &
  \multicolumn{2}{c}{\textbf{With RAG}} \\
\cmidrule(lr){3-5} \cmidrule(lr){6-7}
 & & \textbf{Best Result} & \textbf{Prompt} & \textbf{Model} & \textbf{Best Result} & \textbf{Model} \\
\midrule

\multirow{7}{*}{BI-RADS}
 & Accuracy    & 0.57          & CoT     & Qwen2.5VL & \textbf{0.895}   & MedGemma \\
 & Precision   & 0.8139          & Fewshot     & LLAVA-Med & \textbf{0.9757}  & MedGemma \\
 & Recall      & 0.57          & CoT     & Qwen2.5VL & \textbf{0.895}   & MedGemma \\
 & F1-score    & 0.59          & CoT     & Qwen2.5VL & \textbf{0.93002} & MedGemma \\
 & Specificity & 0.8602          & Fewshot     & LLAVA-Med & \textbf{0.9807}  & MedGemma \\
 & BERTScore   & \textbf{0.9999}          & Fewshot & Qwen2.5VL & \textbf{0.9999}  & MedGemma \\
 & ROUGE-L     & \textbf{0.4906} & Fewshot & MedGemma  & \textbf{0.4906}           & MedGemma \\
\midrule

\multirow{7}{*}{Breast Density}
 & Accuracy    & 0.3589          & Fewshot & MedGemma  & \textbf{0.3739} & MedGemma \\
 & Precision   & 0.4585          & CoT     & Qwen2.5VL & \textbf{0.65}   & MedGemma \\
 & Recall      & 0.3589          & Fewshot & MedGemma  & \textbf{0.3739} & MedGemma \\
 & F1-score    & \textbf{0.3715} & Fewshot & MedGemma  & 0.3605          & MedGemma \\
 & Specificity & \textbf{0.7886} & Fewshot & MedGemma  & 0.7877          & MedGemma \\
 & BERTScore   & 0.886           & Fewshot & MedGemma  & \textbf{0.8991} & MedGemma \\
 & ROUGE-L     & 0.3964          & Fewshot & MedGemma  & \textbf{0.4939} & MedGemma \\
\midrule

\multirow{2}{*}{Findings}
 & BERTScore   & \textbf{0.9017} & Fewshot & MedGemma  & 0.8615          & LLAVA-Med \\
 & ROUGE-L     & \textbf{0.5423} & Fewshot & MedGemma  & 0.2706          & LLAVA-Med \\
\bottomrule
\end{tabular}%
}
\end{table*}

Detailed results are provided in the Appendix. Here, we focus on the key findings relevant to our research questions.
\subsection{Report generation (textual similarity)}
We first evaluate report-generation similarity using only text-similarity metrics (BERTScore and ROUGE-L) on narrative fields, BI-RADS, density, and findings directly read by clinicians. On VinDr-Mammo (Table \ref{new_vindr_with_rag}), retrieval generally improves narrative similarity beyond prompting alone. The best BI-RADS-text BERTScore increases from 0.9313 (few-shot, \textit{Qwen2.5-VL}) to 0.9401 (RAG,\textit{ MedGemma}), while ROUGE-L improves from 0.3593 to 0.4347 (RAG, \textit{MedGemma}), indicating better lexical alignment and phrasing consistency for the clinically salient BI-RADS narrative. For density text, ROUGE-L improves from 0.8643 (few-shot, \textit{Qwen2.5-VL}) to 0.8740 (RAG, \textit{Qwen2.5-VL}), and for the findings text, ROUGE-L increases from 0.4090 (few-shot, \textit{LLaVA-Med}) to 0.4652 (RAG, \textit{Qwen2.5-VL}). Overall, on VinDr-Mammo, RAG yields small but consistent gains in narrative similarity, suggesting that retrieved context helps the model better anchor phrasing and content for structured report fields expressed in free text.

On DMID (Table \ref{new_dmid_with_rag}), narrative trends are more mixed and highlight that retrieval quality and corpus fit can strongly affect textual similarity. BI-RADS, BERTScore, and ROUGE-L remain the same. Density text improves more clearly, with BERTScore increasing from 0.8860 to 0.8991 and ROUGE-L from 0.3964 to 0.4939. In contrast, the similarity of the findings' text drops substantially under RAG (BERTScore from 0.9017 to 0.8615 and ROUGE-L from 0.5423 to 0.2706). These results suggest that retrieval can reinforce narrative similarity (as in density) or introduce distracting or mismatched context (as in findings), depending on how well the retrieved examples align with the target distribution and the constraints of the prompt. These DMID results illustrate that RAG is not universally beneficial for narrative similarity. The DMID reports are short and already tightly paired with each image. In this setting, retrieval can introduce redundant or mismatched context, leading to lexical and semantic noise and reducing overlap-based metrics even when the model remains clinically plausible. In particular, retrieved examples may differ in the type, distribution, or phrasing conventions of the abnormality, causing the generated findings sentence to drift toward the retrieved style rather than the ground-truth wording. Moreover, when multiple retrieved examples are injected, the model may over-condition on the example text, effectively ``crowding out'' direct image-based reasoning. Overall, RAG effectiveness depends on the image and report, on the retrieval, and on how strongly prompts constrain the model to prioritize the input image over retrieved narratives.

\subsection{Prompting vs RAG for structured labels}
We next compare prompting and RAG for structured label prediction without fine-tuning, using classification metrics for multi-class tasks (BI-RADS, density) and, where available, binary findings (calcification, mass, asymmetry, suspicion). On VinDr-Mammo (Table \ref{new_vindr_with_rag}), RAG tends to provide modest improvements in structured prediction compared to best-prompt baselines. For BI-RADS classification, accuracy improves from 0.2909 (few-shot, \textit{LLaVA-Med}) to 0.3190 (RAG, \textit{MedGemma}), with corresponding improvements in F1-score from 0.2217 to 0.2756 (RAG, \textit{Qwen2.5-VL}). Breast density accuracy improves from 0.7832 to 0.8068 (both best with \textit{Qwen2.5-VL}), and calcification shows stronger gains in several metrics (accuracy 0.8391 to 0.8591 and F1-score 0.7679 to 0.8432). The Mass prediction improves moderately (accuracy 0.6218 to 0.6582 and F1-score 0.5016 to 0.5726). The asymmetry illustrates the metric-dependent behavior. That is, accuracy decreases from 0.8915 (zero-shot, \textit{LLaVA-Med}) to 0.8664 (RAG, \textit{Qwen2.5-VL}), yet the F1-score improves from 0.7875 to 0.8373, consistent with RAG shifting the operating point rather than uniformly improving all metrics. Suspicion prediction improves (accuracy 0.6841 to 0.7350 and F1-score 0.5897 to 0.6961). In aggregate, RAG typically yields incremental gains for structured labels, but improvements are not uniformly monotonic across tasks or metrics.

On DMID (Table \ref{new_dmid_with_rag}), the effect of RAG on structured labels is dataset-dependent and can be much larger. For BI-RADS classification, RAG produces a substantial increase in accuracy (0.57 to 0.8950) and F1-score (0.59 to 0.9300), suggesting that retrieved examples can dramatically stabilize label inference in this setting. However, density classification shows only slight changes in accuracy (0.3589 to 0.3739) and decreases in F1-score (0.3715 to 0.3605), suggesting that retrieval gains are not guaranteed and may depend on label distribution, report style, and the degree to which retrieval provides discriminative cues for the target label.

\vspace{0.15cm}
\begin{mdframed}[backgroundcolor=blue!10, linecolor=black!60, roundcorner=10pt]
\textbf{RQ1: How well do local medical VLMs generate structured reports under prompting vs RAG?}
\\ \\
\textbf{$\textbf{RQ}_{\textcolor{red}{1}}$ Findings:} 
Local medical VLMs can generate high-quality structured reports under prompting alone, particularly with few-shot prompting. Adding RAG generally improves report similarity and often improves downstream label quality, with noticeable narrative gains. However, the benefit is not uniform across all text fields, indicating sensitivity to dataset characteristics and retrieval fit.
\end{mdframed}

% Please add the following required packages to your document preamble:
% \usepackage{multirow}
% \usepackage{graphicx}
% \usepackage{booktabs}

\begin{table*}[ht]
\centering
\caption{Performance comparison of MedGemma model}
\label{finetune}
\resizebox{\textwidth}{!}{%
\begin{tabular}{llccccccccccc}
\toprule
\multirow{3}{*}{\textbf{Task}} &
  \multirow{3}{*}{\textbf{Evaluation Parameter}} &
  \multirow{3}{*}{\textbf{Zeroshot}} &
  \multirow{3}{*}{\textbf{Fewshot}} &
  \multirow{3}{*}{\textbf{CoT}} &
  \multirow{3}{*}{\textbf{RAG-Fewshot}} &
  \multicolumn{6}{c}{\textbf{Finetune}} \\
\cmidrule(lr){7-12}
 & & & & & &
  \multicolumn{4}{c}{\textbf{Generate All Tasks Together}} &
  \multicolumn{2}{c}{\textbf{Generate 1 Task at a Time}} \\
\cmidrule(lr){7-10} \cmidrule(lr){11-12}
 & & & & & &
  \textbf{3 Epochs} & \textbf{6 Epochs} & \textbf{10 Epochs} & \textbf{15 Epochs} &
  \textbf{6 Epochs} & \textbf{10 Epochs} \\
\midrule

\multirow{5}{*}{BI-RADS}
 & Accuracy    & 0.2218 & 0.1019 & 0.2493 & 0.319  & 0.3564 & 0.4698 & 0.6355          & 0.5139 & 0.4976 & \textbf{0.7545} \\
 & Precision   & 0.2722 & 0.1515 & 0.1846 & 0.3663 & 0.3608 & 0.5734 & 0.6544          & 0.5444 & 0.484  & \textbf{0.7356} \\
 & Recall      & 0.2218 & 0.1019 & 0.2493 & 0.319  & 0.3564 & 0.4698 & 0.6355          & 0.5139 & 0.4591 & \textbf{0.7516} \\
 & F1-score    & 0.1237 & 0.0376 & 0.1412 & 0.2698 & 0.3489 & 0.4186 & 0.6261          & 0.4911 & 0.4152 & \textbf{0.7404} \\
 & Specificity & 0.8006 & 0.8004 & 0.8059 & 0.8265 & 0.8353 & 0.8629 & 0.906           & 0.875  & 0.8612 & \textbf{0.9387} \\
\midrule

\multirow{5}{*}{Breast Density}
 & Accuracy    & 0.7173 & 0.5921 & 0.742  & 0.7862 & 0.7736 & 0.8314 & \textbf{0.884}  & 0.7922 & 0.5606 & 0.8262 \\
 & Precision   & 0.6507 & 0.6818 & 0.6631 & 0.7472 & 0.7606 & 0.5826 & \textbf{0.8967} & 0.8771 & 0.6018 & 0.8831 \\
 & Recall      & 0.7173 & 0.5921 & 0.742  & 0.7862 & 0.7736 & 0.8485 & \textbf{0.884}  & 0.7922 & 0.5606 & 0.8677 \\
 & F1-score    & 0.6819 & 0.6061 & 0.69   & 0.7529 & 0.7654 & 0.6147 & \textbf{0.8894} & 0.8145 & 0.5172 & 0.873  \\
 & Specificity & 0.7754 & 0.7649 & 0.7667 & 0.7998 & 0.8224 & 0.9463 & \textbf{0.9312} & 0.9294 & 0.8764 & 0.9271 \\
\midrule

\multirow{5}{*}{Calcification}
 & Accuracy    & 0.4009 & 0.2055 & 0.6641 & 0.7441 & 0.8164 & 0.8623 & \textbf{0.9341} & 0.9186 & 0.8388 & 0.7621 \\
 & Precision   & 0.7461 & 0.7725 & 0.7466 & 0.8162 & 0.8421 & 0.5534 & \textbf{0.9317} & 0.9179 & 0.9425 & 0.9257 \\
 & Recall      & 0.4009 & 0.2055 & 0.6641 & 0.7441 & 0.8164 & 0.8078 & \textbf{0.9341} & 0.9186 & 0.9423 & 0.9209 \\
 & F1-score    & 0.4535 & 0.1368 & 0.6973 & 0.7694 & 0.8266 & 0.6569 & \textbf{0.9313} & 0.9182 & 0.9384 & 0.9107 \\
 & Specificity & 0.5254 & 0.5107 & 0.5414 & 0.6879 & 0.7322 & \textbf{0.8729} & 0.8418          & 0.846  & 0.8388 & 0.7621 \\
\midrule

\multirow{5}{*}{Mass}
 & Accuracy    & 0.4945 & 0.4127 & 0.4914 & 0.5441 & 0.595  & 0.7791 & 0.7541 & 0.7586 & 0.8455 & \textbf{0.874}  \\
 & Precision   & 0.5272 & 0.6556 & 0.5305 & 0.6221 & 0.6173 & 0.6964 & 0.7682 & 0.7555 & 0.8574 & \textbf{0.8787} \\
 & Recall      & 0.4945 & 0.4127 & 0.4914 & 0.5441 & 0.595  & 0.7446 & 0.7541 & 0.7586 & 0.8582 & \textbf{0.8773} \\
 & F1-score    & 0.5016 & 0.2885 & 0.4976 & 0.5404 & 0.6007 & 0.7197 & 0.7571 & 0.7519 & 0.8576 & \textbf{0.8777} \\
 & Specificity & 0.4988 & 0.5211 & 0.5022 & 0.5845 & 0.5951 & 0.8003 & 0.7578 & 0.724  & 0.8455 & \textbf{0.874}  \\
\midrule

\multirow{5}{*}{Asymmetry}
 & Accuracy    & 0.8545 & 0.8259 & 0.8545 & 0.7868 & 0.7882 & 0.8823          & \textbf{0.8605} & 0.8423 & 0.8536 & 0.8536 \\
 & Precision   & 0.7302 & 0.7359 & 0.7302 & 0.7713 & 0.7854 & 0.5631          & \textbf{0.8292} & 0.8042 & 0.7287 & 0.7287 \\
 & Recall      & 0.8545 & 0.8259 & 0.8545 & 0.7868 & 0.7882 & 0.85            & \textbf{0.8605} & 0.8423 & 0.8536 & 0.8536 \\
 & F1-score    & 0.7875 & 0.7759 & 0.7875 & 0.7787 & 0.7868 & 0.6775          & \textbf{0.8261} & 0.8152 & 0.7862 & 0.7862 \\
 & Specificity & 0.5    & 0.4884 & 0.5    & 0.5369 & 0.5675 & 0.8878          & \textbf{0.5722} & 0.5667 & 0.5    & 0.5    \\
\midrule

\multirow{5}{*}{Suspicion}
 & Accuracy    & 0.3464 & 0.3291 & 0.3982 & 0.4134 & 0.6735 & 0.7495 & \textbf{0.7981} & 0.789  & 0.8536 & 0.7146 \\
 & Precision   & 0.6325 & 0.7579 & 0.6464 & 0.5853 & 0.7126 & 0.8842 & \textbf{0.7939} & 0.7555 & 0.6307 & 0.7179 \\
 & Recall      & 0.3464 & 0.3291 & 0.3982 & 0.4134 & 0.6735 & 0.7223 & \textbf{0.7981} & 0.789  & 0.6173 & 0.6868 \\
 & F1-score    & 0.2277 & 0.1771 & 0.3369 & 0.3271 & 0.6836 & 0.7951 & \textbf{0.7865} & 0.7519 & 0.6179 & 0.6874 \\
 & Specificity & 0.5111 & 0.5076 & 0.5347 & 0.5174 & 0.6741 & 0.8056 & \textbf{0.7263} & 0.724  & 0.8061 & 0.8478 \\
\bottomrule
\end{tabular}%
}
\end{table*}

\subsection{Effect of QLoRA fine-tuning}
Finally, we show the impact of PEFT on MedGemma using QLoRA (Table \ref{finetune}). In addition to standard prompting and RAG, Table~4 separates two fine-tuning output formats: \textit{multi-task generation}, where the model produces the full JSON report with all fields in one response, and \textit{single-task generation}, where the model is prompted to generate only one task/field at a time. This distinction matters because single-task decoding can reduce output coupling across fields, while multi-task decoding preserves the intended end-to-end report setting.

In general, QLoRA produces substantially higher gains than prompting or RAG in structured label prediction, supporting an "escalating reliability ladder” for classification: prompting provides a usable baseline, RAG can stabilize performance in some tasks, and fine-tuning produces the most consistent improvements when reliable label accuracy is required. For example, BI-RADS accuracy improves from the best prompting setting (0.2493) to 0.3190 with RAG, and further to 0.6355 with multi-task QLoRA at 10 epochs; notably, single-task QLoRA achieves an even higher BI-RADS accuracy of 0.7545 at 10 epochs. Breast density prediction improves similarly from 0.7420 (best prompt) to 0.7862 (RAG) and to 0.8840 with multi-task QLoRA at 10 epochs. Calcification prediction improves from 0.6641 (best prompt) to 0.7441 (RAG) and to 0.9341 with multi-task QLoRA at 10 epochs. Mass accuracy increases from 0.5441 (RAG) to 0.7791 under multi-task QLoRA at 6 epochs, and rises further under single-task QLoRA to 0.8740 at 10 epochs. Suspicion (derived from BI-RADS) improves from 0.3982 (best prompt) to 0.4134 (RAG) and to 0.7981 with multi-task QLoRA at 10 epochs.

Fine-tuning effects are not always monotonic with additional epochs, and the optimal checkpoint can differ across tasks and output formats. For instance, in multi-task mode, BI-RADS peaks at 10 epochs (0.6355) but drops at 15 epochs (0.5139), and mass peaks at 6 epochs (0.7791) with only minor variation thereafter. Asymmetry highlights an additional nuance: RAG reduces accuracy relative to the best prompt (0.8545 to 0.7868), but multi-task QLoRA recovers and peaks at 0.8823 at 6 epochs; in contrast, single-task asymmetry accuracy remains the same (0.8536 for both 6 and 10 epochs), suggesting that single-task fine-tuning may be ineffective for certain labels in this setup. Taken together, these results indicate that QLoRA generally increases reliability for structured labels, but optimal training duration and even the preferred output format (multi-task vs single-task) are task-dependent and may require checkpoint selection or early stopping.

% \begin{lstlisting}[]
% Multi-Task Generation:

% Input: <Prompt + Image>
% Output:
% {
%   "image_id": "image_file_path_1",
%   "breast_density": "Density C - Heterogeneously Dense. More of the breast is made of dense glandular and fibrous tissue. This can make it hard to see small masses in or around the dense tissue, which also appear as white areas.",
%   "BI-RADS": "BI-RADS 1 - Negative. Healthy Breast.",
%   "findings": "Healthy Breast. No Findings",
%   "suspicion": "healthy"
% }
% \end{lstlisting}

% \begin{lstlisting}[]
% Single-Task Generation:

% Input: <Prompt + Image>
% Output:
% {
%   "image_id": "image_file_path_1",
%   "BI-RADS": "BI-RADS 1 - Negative. Healthy Breast.",
% }
% \end{lstlisting}

Summary of macro F1 gains from RAG and fine-tuning. Table \ref{tab:delta_ladder_f1} summarizes the F1 gains from adding RAG and QLoRA fine-tuning for the \textit{MedGemma} model on the VinDr-Mammo dataset. We use the macro F1-score, which better reflects minority-class performance in imbalanced settings. Best Prompt is the maximum over zero-shot, few-shot, and CoT prompting.
Best RAG is the maximum over the corresponding retrieval-augmented settings. Best QLoRA selects the best-performing fine-tuned checkpoint across both output formats, multi-task full-JSON generation, and single-task one-field generation, because the optimal format/checkpoint can be task-dependent. The negative gains (asymmetry and suspicion) indicate that retrieval can occasionally introduce context mismatches, whereas QLoRA yields the strongest and most consistent gains across tasks.
Only the F1-score is compared here, as it is the harmonic mean of precision and recall \cite{jahangir2025ecg, jahangir2023proposing}.

% Please add the following required packages to your document preamble:
% \usepackage{booktabs}

\begin{table*}[!ht]
\centering
\small
\caption{Gain summary of adding RAG and fine-tuning for MedGemma on VinDr-Mammo (macro F1-score).}
\label{tab:delta_ladder_f1}
\begin{tabular}{lccccc}
\toprule
\textbf{Task} & \textbf{Best Prompt} & \textbf{Best RAG} & $\boldsymbol{\Delta}$\textbf{(RAG\,--\,Prompt)} & \textbf{Best QLoRA} & $\boldsymbol{\Delta}$\textbf{(QLoRA\,--\,RAG)} \\
\midrule
BI-RADS        & 0.1412 & 0.2698 & $+$0.1286 & 0.7404 & $+$0.4706 \\
Density       & 0.6900 & 0.7529 & $+$0.0629 & 0.8894 & $+$0.1365 \\
Calcification & 0.6973 & 0.7694 & $+$0.0721 & 0.9384 & $+$0.1690 \\
Mass          & 0.5016 & 0.5404 & $+$0.0388 & 0.8777 & $+$0.3373 \\
Asymmetry     & 0.7875 & 0.7787 & $-$0.0088 & 0.8261 & $+$0.0474 \\
Suspicion     & 0.3369 & 0.3271 & $-$0.0098 & 0.7951 & $+$0.4680 \\
\bottomrule
\end{tabular}
\end{table*}

\vspace{0.15cm}
\begin{mdframed}[backgroundcolor=blue!10, linecolor=black!60, roundcorner=10pt]
\textbf{RQ2: When does lightweight fine-tuning outperform prompting/RAG for classification?}
\\ \\
\textbf{$\textbf{RQ}_{\textcolor{red}{2}}$ Findings:} Lightweight fine-tuning (QLoRA) outperforms prompting/RAG most strongly for hard structured classification. Overall, the evidence supports a practical split: prompting/RAG for strong report drafting and grounding, and fine-tuning when dependable label accuracy is required.
\end{mdframed}

% Please add the following required packages to your document preamble:
% \usepackage{graphicx}
% \usepackage{booktabs}

\begin{table*}[!ht]
\centering
\caption{Comparison of classification accuracy on several tasks of \textit{MammoWise} models with other SOTA}
\label{comparison}
\resizebox{\textwidth}{!}{%
\begin{tabular}{lccccc}
\toprule
\textbf{Task} & \textbf{Our Work} & \textbf{MammoCLIP} & \textbf{LLaVA-Mammo} & \textbf{LLaVA-MultiMammo} & \textbf{PubMedCLIP} \\
\midrule
BI-RADS         & \textbf{0.7545} & --             & --    & --    & 0.5325 \\
Breast Density & \textbf{0.884}  & 0.88           & 0.766  & 0.806  & --     \\
Mass           & \textbf{0.8740} & 0.8            & --    & --    & --     \\
Calcification  & 0.9341          & \textbf{0.98}  & --    & --    & --     \\
\bottomrule
\end{tabular}%
}
\end{table*}

To compare to existing approaches, we use MammoWise to compare our fine-tuned \textit{MedGemma} model against reported SOTA results in the literature. Because papers often report results across different task subsets and with diverse metrics, a one-to-one comparison for every task is not always possible. Table \ref{comparison} summarizes key comparisons of accuracy, where available.

On VinDr-Mammo, \textit{PubMedCLIP} achieves 0.5325 accuracy for BI-RADS classification, while our fine-tuned \textit{MedGemma} achieves 0.7545 with single-task training, an improvement on the same dataset and metric. For breast-density classification, \textit{LLaVA-Mammo} and \textit{LLaVA-MultiMammo} report accuracies of 0.766 and 0.806, respectively; our model reaches 0.884. For mass and calcification detection, our fine-tuned model (0.8740) outperforms \textit{MammoCLIP} (0.8) on mass classification but falls short on calcification (0.9341 vs. 0.98).

For report generation, we found no peer-reviewed mammography papers that report end-to-end BERTScore and ROUGE-L metrics for complete narrative mammogram reports, making direct external comparison impossible. Most contemporary VLM work focuses on classification, localization, or visual question answering rather than full, structured reporting. Nonetheless, \textit{MedGemma} achieves strong generation metrics and produces clinically structured JSON reports, offering capabilities that go beyond pure classification.

Overall, \textit{MammoWise} demonstrates that a multi-model pipeline powered by open-source VLMs can be used to effectively iterate over models and to converge on a model that matches or exceeds SOTA baselines on key classification tasks, while also enabling high-quality report generation that has been underexplored in prior work.

Our study has several limitations, mainly due to computational and scope constraints. First, we did not explore more advanced prompting strategies such as Tree-of-Thought \cite{yao2023tree}, or Reason+Act prompting \cite{yao2023react}. Second, we tested only three open-source VLMs; other models may offer different trade-offs in performance and efficiency. Third, we did not fine-tune larger models that might further improve performance but require more powerful hardware. Fourth, we did not investigate hybrid architectures that combine specialist mammography encoders with general-purpose decoders. Fifth, because only two datasets were used, there may be a domain shift (e.g., scale, label noise), leading to different results between the VinDr-Mammo and DMID datasets. Additionally, our QLoRA fine-tuning uses a class-rebalanced subset (Table \ref{tab:BI-RADS_balance}) constructed via downsampling and augmentation, which differs from the original test distribution. This distribution shift can inflate macro-averaged metrics while potentially harming calibration or operating characteristics under real-world prevalence. We therefore interpret fine-tuning gains primarily as evidence of improved \emph{task reliability under balanced emphasis}, rather than a definitive statement about screening-time calibration. Finally, we evaluated only single-exam images. Thus, extending MammoWise to longitudinal mammograms and richer multimodal clinical data remains an important direction for future work.

\section{Conclusion}
In this study, we present \textit{MammoWise}, a novel multi-model pipeline that combines open-source medical VLMs, tailored prompting techniques, and retrieval-augmented generation to produce structured mammogram reports and perform key classification tasks. Our experiments showed that while base prompting strategies yield variable classification performance, they already support strong report generation, particularly when combined with few-shot prompting. RAG-based prompting further improves text-generation quality and provides context-driven grounding. Parameter-efficient fine-tuning of \textit{MedGemma} significantly increases classification performance, especially for BI-RADS, breast density, and calcification, demonstrating that local, resource-aware adaptation can close much of the gap to specialized SOTA models.

Overall, \textit{MammoWise} offers a practical blueprint for deploying local VLMs as flexible, privacy-preserving assistants in breast cancer screening. By decoupling the pipeline from any single model and supporting prompting, RAG, and fine-tuning within one framework, it creates a reusable platform for future research and clinical prototyping in mammography and beyond.

\section*{Data \& Code Availability Statement}
The code for this research and the pipeline tool are available on \href{https://github.com/RaiyanJahangir/MammoWise}{GitHub \newline(https://github.com/RaiyanJahangir/MammoWise)}.

\section*{Declaration of Interests}
The authors declare that they have no known competing financial interests or personal
relationships that could have appeared to influence the work reported in this paper.

% \section*{Acknowledgement}
\bibliographystyle{unsrt}
\bibliography{reference}

@inproceedings{ghosh2024mammo,
  title={Mammo-clip: A vision language foundation model to enhance data efficiency and robustness in mammography},
  author={Ghosh, Shantanu and Poynton, Clare B and Visweswaran, Shyam and Batmanghelich, Kayhan},
  booktitle={International Conference on Medical Image Computing and Computer-Assisted Intervention},
  pages={632--642},
  year={2024},
  organization={Springer}
}

@inproceedings{jain2024mmbcd,
  title={MMBCD: Multimodal Breast Cancer Detection from Mammograms with Clinical History},
  author={Jain, Kshitiz and Bansal, Aditya and Rangarajan, Krithika and Arora, Chetan},
  booktitle={International Conference on Medical Image Computing and Computer-Assisted Intervention},
  pages={144--154},
  year={2024},
  organization={Springer}
}

@inproceedings{de2024unlocking,
  title={Unlocking The Potential Of Vision-Language Models For Mammography Analysis},
  author={de Moura, Lu{\'\i}s Vin{\'\i}cius and Ravazio, Rafaela and Mattjie, Christian and Kupssinsk{\"u}, Lucas Silveira and Freitas, Carla Maria Dal Sasso and Barros, Rodrigo C},
  booktitle={2024 IEEE International Symposium on Biomedical Imaging (ISBI)},
  pages={1--4},
  year={2024},
  organization={IEEE}
}

@inproceedings{urooj2024knowledge,
  title={Knowledge-Grounded Adaptation Strategy for Vision-Language Models: Building a Unique Case-Set for Screening Mammograms for Residents Training},
  author={Urooj Khan, Aisha and Garrett, John and Bradshaw, Tyler and Salkowski, Lonie and Jeong, Jiwoong and Tariq, Amara and Banerjee, Imon},
  booktitle={International Conference on Medical Image Computing and Computer-Assisted Intervention},
  pages={587--598},
  year={2024},
  organization={Springer}
}

@article{lashof2001mammography,
  title={Mammography and beyond: developing technologies for the early detection of breast cancer},
  author={Lashof, Joyce C and Henderson, I Craig and Nass, Sharyl J},
  year={2001},
  publisher={National Academies Press}
}

@inproceedings{jia2022visual,
  title={Visual prompt tuning},
  author={Jia, Menglin and Tang, Luming and Chen, Bor-Chun and Cardie, Claire and Belongie, Serge and Hariharan, Bharath and Lim, Ser-Nam},
  booktitle={European Conference on Computer Vision},
  pages={709--727},
  year={2022},
  organization={Springer}
}

@article{oza2024digital,
  title={Digital mammography dataset for breast cancer diagnosis research (DMID) with breast mass segmentation analysis},
  author={Oza, Parita and Oza, Urvi and Oza, Rajiv and Sharma, Paawan and Patel, Samir and Kumar, Pankaj and Gohel, Bakul},
  journal={Biomedical Engineering Letters},
  volume={14},
  number={2},
  pages={317--330},
  year={2024},
  publisher={Springer}
}

@article{nguyen2023vindr,
  title={VinDr-Mammo: A large-scale benchmark dataset for computer-aided diagnosis in full-field digital mammography},
  author={Nguyen, Hieu T and Nguyen, Ha Q and Pham, Hieu H and Lam, Khanh and Le, Linh T and Dao, Minh and Vu, Van},
  journal={Scientific Data},
  volume={10},
  number={1},
  pages={277},
  year={2023},
  publisher={Nature Publishing Group UK London}
}

@article{zhang2024vision,
  title={Vision-language models for vision tasks: A survey},
  author={Zhang, Jingyi and Huang, Jiaxing and Jin, Sheng and Lu, Shijian},
  journal={IEEE Transactions on Pattern Analysis and Machine Intelligence},
  year={2024},
  publisher={IEEE}
}

@article{ji2023survey,
  title={Survey of hallucination in natural language generation},
  author={Ji, Ziwei and Lee, Nayeon and Frieske, Rita and Yu, Tiezheng and Su, Dan and Xu, Yan and Ishii, Etsuko and Bang, Ye Jin and Madotto, Andrea and Fung, Pascale},
  journal={ACM computing surveys},
  volume={55},
  number={12},
  pages={1--38},
  year={2023},
  publisher={ACM New York, NY}
}

@article{cao2025mammovlm,
  title={MammoVLM: A generative large vision-language model for mammography-related diagnostic assistance},
  author={Cao, Zhenjie and Deng, Zhuo and Ma, Jie and Hu, Jintao and Ma, Lan},
  journal={Information Fusion},
  pages={102998},
  year={2025},
  publisher={Elsevier}
}

@article{liu2023visual,
  title={Visual instruction tuning},
  author={Liu, Haotian and Li, Chunyuan and Wu, Qingyang and Lee, Yong Jae},
  journal={Advances in neural information processing systems},
  volume={36},
  pages={34892--34916},
  year={2023}
}

@article{bai2025qwen2,
  title={Qwen2. 5-vl technical report},
  author={Bai, Shuai and Chen, Keqin and Liu, Xuejing and Wang, Jialin and Ge, Wenbin and Song, Sibo and Dang, Kai and Wang, Peng and Wang, Shijie and Tang, Jun and others},
  journal={arXiv preprint arXiv:2502.13923},
  year={2025}
}

@article{touvron2023llama,
  title={Llama: Open and efficient foundation language models},
  author={Touvron, Hugo and Lavril, Thibaut and Izacard, Gautier and Martinet, Xavier and Lachaux, Marie-Anne and Lacroix, Timoth{\'e}e and Rozi{\`e}re, Baptiste and Goyal, Naman and Hambro, Eric and Azhar, Faisal and others},
  journal={arXiv preprint arXiv:2302.13971},
  year={2023}
}

@inproceedings{zamfirescu2023johnny,
  title={Why Johnny can’t prompt: how non-AI experts try (and fail) to design LLM prompts},
  author={Zamfirescu-Pereira, J Diego and Wong, Richmond Y and Hartmann, Bjoern and Yang, Qian},
  booktitle={Proceedings of the 2023 CHI conference on human factors in computing systems},
  pages={1--21},
  year={2023}
}

@article{pourpanah2022review,
  title={A review of generalized zero-shot learning methods},
  author={Pourpanah, Farhad and Abdar, Moloud and Luo, Yuxuan and Zhou, Xinlei and Wang, Ran and Lim, Chee Peng and Wang, Xi-Zhao and Wu, QM Jonathan},
  journal={IEEE transactions on pattern analysis and machine intelligence},
  volume={45},
  number={4},
  pages={4051--4070},
  year={2022},
  publisher={IEEE}
}

@article{wang2020generalizing,
  title={Generalizing from a few examples: A survey on few-shot learning},
  author={Wang, Yaqing and Yao, Quanming and Kwok, James T and Ni, Lionel M},
  journal={ACM computing surveys (csur)},
  volume={53},
  number={3},
  pages={1--34},
  year={2020},
  publisher={ACM New York, NY, USA}
}

@article{wei2022chain,
  title={Chain-of-thought prompting elicits reasoning in large language models},
  author={Wei, Jason and Wang, Xuezhi and Schuurmans, Dale and Bosma, Maarten and Xia, Fei and Chi, Ed and Le, Quoc V and Zhou, Denny and others},
  journal={Advances in neural information processing systems},
  volume={35},
  pages={24824--24837},
  year={2022}
}

@article{dettmers2023qlora,
  title={Qlora: Efficient finetuning of quantized llms},
  author={Dettmers, Tim and Pagnoni, Artidoro and Holtzman, Ari and Zettlemoyer, Luke},
  journal={Advances in neural information processing systems},
  volume={36},
  pages={10088--10115},
  year={2023}
}

@inproceedings{renze2024effect,
  title={The effect of sampling temperature on problem solving in large language models},
  author={Renze, Matthew},
  booktitle={Findings of the Association for Computational Linguistics: EMNLP 2024},
  pages={7346--7356},
  year={2024}
}

@article{zhang2019bertscore,
  title={Bertscore: Evaluating text generation with bert},
  author={Zhang, Tianyi and Kishore, Varsha and Wu, Felix and Weinberger, Kilian Q and Artzi, Yoav},
  journal={arXiv preprint arXiv:1904.09675},
  year={2019}
}

@article{yao2023tree,
  title={Tree of thoughts: Deliberate problem solving with large language models},
  author={Yao, Shunyu and Yu, Dian and Zhao, Jeffrey and Shafran, Izhak and Griffiths, Tom and Cao, Yuan and Narasimhan, Karthik},
  journal={Advances in neural information processing systems},
  volume={36},
  pages={11809--11822},
  year={2023}
}

@inproceedings{yao2023react,
  title={React: Synergizing reasoning and acting in language models},
  author={Yao, Shunyu and Zhao, Jeffrey and Yu, Dian and Du, Nan and Shafran, Izhak and Narasimhan, Karthik and Cao, Yuan},
  booktitle={International Conference on Learning Representations (ICLR)},
  year={2023}
}

@inproceedings{jahangir2023comparative,
  title={A Comparative Analysis of Potato Leaf Disease Classification with Big Transfer (BiT) and Vision Transformer (ViT) Models},
  author={Jahangir, Raiyan and Sakib, Tanjim and Baki, Ramisha and Hossain, Md Mushfique},
  booktitle={2023 IEEE 9th International Women in Engineering (WIE) Conference on Electrical and Computer Engineering (WIECON-ECE)},
  pages={58--63},
  year={2023},
  organization={IEEE}
}

@article{jahangir2025ecg,
  title={ECG-based heart arrhythmia classification using feature engineering and a hybrid stacked machine learning},
  author={Jahangir, Raiyan and Islam, Muhammad Nazrul and Islam, Md Shofiqul and Islam, Md Motaharul},
  journal={BMC Cardiovascular Disorders},
  volume={25},
  number={1},
  pages={260},
  year={2025},
  publisher={Springer}
}

@inproceedings{jahangir2023performance,
  title={A performance analysis of brain tumor classification from MRI images using vision transformers and CNN-based classifiers},
  author={Jahangir, Raiyan and Sakib, Tanjim and Haque, Riasat and Kamal, Mahedi},
  booktitle={2023 26th International Conference on Computer and Information Technology (ICCIT)},
  pages={1--6},
  year={2023},
  organization={IEEE}
}

@article{raiaan2024review,
  title={A review on large language models: Architectures, applications, taxonomies, open issues and challenges},
  author={Raiaan, Mohaimenul Azam Khan and Mukta, Md Saddam Hossain and Fatema, Kaniz and Fahad, Nur Mohammad and Sakib, Sadman and Mim, Most Marufatul Jannat and Ahmad, Jubaer and Ali, Mohammed Eunus and Azam, Sami},
  journal={IEEE access},
  volume={12},
  pages={26839--26874},
  year={2024},
  publisher={IEEE}
}

@article{sellergren2025medgemma,
  title={MedGemma Technical Report},
  author={Sellergren, Andrew and Kazemzadeh, Sahar and Jaroensri, Tiam and Kiraly, Atilla and Traverse, Madeleine and Kohlberger, Timo and Xu, Shawn and Jamil, Fayaz and Hughes, C{\'\i}an and Lau, Charles and others},
  journal={arXiv preprint arXiv:2507.05201},
  year={2025}
}

@article{team2025gemma,
  title={Gemma 3 technical report},
  author={Team, Gemma and Kamath, Aishwarya and Ferret, Johan and Pathak, Shreya and Vieillard, Nino and Merhej, Ramona and Perrin, Sarah and Matejovicova, Tatiana and Ram{\'e}, Alexandre and Rivi{\`e}re, Morgane and others},
  journal={arXiv preprint arXiv:2503.19786},
  year={2025}
}

@inproceedings{sasikala2019breast,
  title={Breast cancer detection based on medio-lateral obliqueview and cranio-caudal view mammograms: an overview},
  author={Sasikala, S and Bharathi, M and Ezhilarasi, M and Arunkumar, S},
  booktitle={2019 IEEE 10th International Conference on Awareness Science and Technology (iCAST)},
  pages={1--6},
  year={2019},
  organization={IEEE}
}

@book{chotai2020breast,
  title={Breast Imaging Essentials},
  author={Chotai, Niketa and Kulkarni, Supriya},
  year={2020},
  publisher={Springer}
}

@article{rawte2023survey,
  title={A survey of hallucination in large foundation models},
  author={Rawte, Vipula and Sheth, Amit and Das, Amitava},
  journal={arXiv preprint arXiv:2309.05922},
  year={2023}
}

@article{haver2024use,
  title={Use of ChatGPT to assign BI-RADS assessment categories to breast imaging reports},
  author={Haver, Hana L and Yi, Paul H and Jeudy, Jean and Bahl, Manisha},
  journal={American Journal of Roentgenology},
  volume={223},
  number={3},
  pages={e2431093},
  year={2024},
  publisher={American Roentgen Ray Society}
}

@article{pesapane2025preliminary,
  title={A preliminary investigation into the potential, pitfalls, and limitations of large language models for mammography interpretation},
  author={Pesapane, Filippo and Nicosia, Luca and Rotili, Anna and Penco, Silvia and Dominelli, Valeria and Trentin, Chiara and Ferrari, Federica and Signorelli, Giulia and Carriero, Serena and Cassano, Enrico},
  journal={Discover Oncology},
  volume={16},
  number={1},
  pages={233},
  year={2025},
  publisher={Springer}
}

@inproceedings{chen2025llava,
  title={LLaVA-Mammo: adapting LLaVA for interactive and interpretable breast cancer assessment},
  author={Chen, Xuxin and Chen, Jingchu and Chen, Xiaoqian and Gichoya, Judy and Trivedi, Hari and Yang, Xiaofeng},
  booktitle={Medical Imaging 2025: Imaging Informatics},
  volume={13411},
  pages={80--88},
  year={2025},
  organization={SPIE}
}

@inproceedings{chen2025llava2,
  title={LLaVA-MultiMammo: adapting vision-language models for explainable and comprehensive multiview mammogram analysis in breast cancer assessment},
  author={Chen, Xuxin and Chen, Jingchu and Chen, Xiaoqian and Gichoya, Judy and Trivedi, Hari and Yang, Xiaofeng},
  booktitle={Medical Imaging 2025: Computer-Aided Diagnosis},
  volume={13407},
  pages={165--173},
  year={2025},
  organization={SPIE}
}

@inproceedings{wang2022medclip,
  title={Medclip: Contrastive learning from unpaired medical images and text},
  author={Wang, Zifeng and Wu, Zhenbang and Agarwal, Dinesh and Sun, Jimeng},
  booktitle={Proceedings of the Conference on Empirical Methods in Natural Language Processing. Conference on Empirical Methods in Natural Language Processing},
  volume={2022},
  pages={3876},
  year={2022}
}

@article{zhang2023biomedclip,
  title={Biomedclip: a multimodal biomedical foundation model pretrained from fifteen million scientific image-text pairs},
  author={Zhang, Sheng and Xu, Yanbo and Usuyama, Naoto and Xu, Hanwen and Bagga, Jaspreet and Tinn, Robert and Preston, Sam and Rao, Rajesh and Wei, Mu and Valluri, Naveen and others},
  journal={arXiv preprint arXiv:2303.00915},
  year={2023}
}

@article{li2025closer,
  title={A closer look at the explainability of Contrastive language-image pre-training},
  author={Li, Yi and Wang, Hualiang and Duan, Yiqun and Zhang, Jiheng and Li, Xiaomeng},
  journal={Pattern Recognition},
  volume={162},
  pages={111409},
  year={2025},
  publisher={Elsevier}
}

@inproceedings{eslami2023pubmedclip,
  title={Pubmedclip: How much does clip benefit visual question answering in the medical domain?},
  author={Eslami, Sedigheh and Meinel, Christoph and De Melo, Gerard},
  booktitle={Findings of the Association for Computational Linguistics: EACL 2023},
  pages={1181--1193},
  year={2023}
}

@article{liu2019roberta,
  title={Roberta: A robustly optimized bert pretraining approach},
  author={Liu, Yinhan and Ott, Myle and Goyal, Naman and Du, Jingfei and Joshi, Mandar and Chen, Danqi and Levy, Omer and Lewis, Mike and Zettlemoyer, Luke and Stoyanov, Veselin},
  journal={arXiv preprint arXiv:1907.11692},
  year={2019}
}

@inproceedings{jahangir2023brain,
  title={Brain Tumor Classification on MRI Images with Big Transfer and Vision Transformer: Comparative Study},
  author={Jahangir, Raiyan and Sakib, Tanjim and Juboraj, Md Fahmid-Ul-Alam and Feroz, Shejuti Binte and Sharar, Md Munkaser Islam},
  booktitle={2023 IEEE 9th International Women in Engineering (WIE) Conference on Electrical and Computer Engineering (WIECON-ECE)},
  pages={46--51},
  year={2023},
  organization={IEEE}
}

@article{lee2020biobert,
  title={BioBERT: a pre-trained biomedical language representation model for biomedical text mining},
  author={Lee, Jinhyuk and Yoon, Wonjin and Kim, Sungdong and Kim, Donghyeon and Kim, Sunkyu and So, Chan Ho and Kang, Jaewoo},
  journal={Bioinformatics},
  volume={36},
  number={4},
  pages={1234--1240},
  year={2020},
  publisher={Oxford University Press}
}

@incollection{marcondes2025using,
  title={Using ollama},
  author={Marcondes, Francisco S and Gala, Adelino and Magalh{\~a}es, Renata and Perez de Britto, Fernando and Dur{\~a}es, Dalila and Novais, Paulo},
  booktitle={Natural Language Analytics with Generative Large-Language Models: A Practical Approach with Ollama and Open-Source LLMs},
  pages={23--35},
  year={2025},
  publisher={Springer}
}

@incollection{jain2022hugging,
  title={Hugging face},
  author={Jain, Shashank Mohan},
  booktitle={Introduction to transformers for NLP: With the hugging face library and models to solve problems},
  pages={51--67},
  year={2022},
  publisher={Springer}
}

@inproceedings{lavanya2024advanced,
  title={Advanced video transcription and summarization A synergy of langchain, language models, and VectorDB with Mozilla deep speech},
  author={Lavanya, K and Aravind, K and Dixit, Vishal and others},
  booktitle={2024 Second International Conference on Emerging Trends in Information Technology and Engineering (ICETITE)},
  pages={1--9},
  year={2024},
  organization={IEEE}
}

@inproceedings{lin2004rouge,
  title={Rouge: A package for automatic evaluation of summaries},
  author={Lin, Chin-Yew},
  booktitle={Text summarization branches out},
  pages={74--81},
  year={2004}
}

@article{sexauer2023diagnostic,
  title={Diagnostic accuracy of automated ACR BI-RADS breast density classification using deep convolutional neural networks},
  author={Sexauer, Raphael and Hejduk, Patryk and Borkowski, Karol and Ruppert, Carlotta and Weikert, Thomas and Dellas, Sophie and Schmidt, Noemi},
  journal={European Radiology},
  volume={33},
  number={7},
  pages={4589--4596},
  year={2023},
  publisher={Springer}
}

@inproceedings{fu2023effectiveness,
  title={On the effectiveness of parameter-efficient fine-tuning},
  author={Fu, Zihao and Yang, Haoran and So, Anthony Man-Cho and Lam, Wai and Bing, Lidong and Collier, Nigel},
  booktitle={Proceedings of the AAAI conference on artificial intelligence},
  volume={37},
  number={11},
  pages={12799--12807},
  year={2023}
}

@inproceedings{sung2022vl,
  title={Vl-adapter: Parameter-efficient transfer learning for vision-and-language tasks},
  author={Sung, Yi-Lin and Cho, Jaemin and Bansal, Mohit},
  booktitle={Proceedings of the IEEE/CVF conference on computer vision and pattern recognition},
  pages={5227--5237},
  year={2022}
}

@article{hu2022lora,
  title={Lora: Low-rank adaptation of large language models.},
  author={Hu, Edward J and Shen, Yelong and Wallis, Phillip and Allen-Zhu, Zeyuan and Li, Yuanzhi and Wang, Shean and Wang, Lu and Chen, Weizhu and others},
  journal={ICLR},
  volume={1},
  number={2},
  pages={3},
  year={2022}
}

@article{li2023llava,
  title={Llava-med: Training a large language-and-vision assistant for biomedicine in one day},
  author={Li, Chunyuan and Wong, Cliff and Zhang, Sheng and Usuyama, Naoto and Liu, Haotian and Yang, Jianwei and Naumann, Tristan and Poon, Hoifung and Gao, Jianfeng},
  journal={Advances in Neural Information Processing Systems},
  volume={36},
  pages={28541--28564},
  year={2023}
}

@article{mesinovic2025explainability,
  title={Explainability in the age of large language models for healthcare},
  author={Mesinovic, Munib and Watkinson, Peter and Zhu, Tingting},
  journal={Communications Engineering},
  volume={4},
  number={1},
  pages={128},
  year={2025},
  publisher={Nature Publishing Group UK London}
}

@article{amann2020explainability,
  title={Explainability for artificial intelligence in healthcare: a multidisciplinary perspective},
  author={Amann, Julia and Blasimme, Alessandro and Vayena, Effy and Frey, Dietmar and Madai, Vince I and Precise4Q Consortium},
  journal={BMC medical informatics and decision making},
  volume={20},
  number={1},
  pages={310},
  year={2020},
  publisher={Springer}
}

@article{patidar2024transparency,
  title={Transparency in AI decision making: A survey of explainable AI methods and applications},
  author={Patidar, Nandkishore and Mishra, Sejal and Jain, Rahul and Prajapati, Dhiren and Solanki, Amit and Suthar, Rajul and Patel, Kavindra and Patel, Hiral},
  journal={Advances of Robotic Technology},
  volume={2},
  number={1},
  year={2024}
}

@article{wang2024qwen2,
  title={Qwen2-vl: Enhancing vision-language model's perception of the world at any resolution},
  author={Wang, Peng and Bai, Shuai and Tan, Sinan and Wang, Shijie and Fan, Zhihao and Bai, Jinze and Chen, Keqin and Liu, Xuejing and Wang, Jialin and Ge, Wenbin and others},
  journal={arXiv preprint arXiv:2409.12191},
  year={2024}
}

@article{giray2023prompt,
  title={Prompt engineering with ChatGPT: a guide for academic writers},
  author={Giray, Louie},
  journal={Annals of biomedical engineering},
  volume={51},
  number={12},
  pages={2629--2633},
  year={2023},
  publisher={Springer}
}

@article{goceri2023medical,
  title={Medical image data augmentation: techniques, comparisons and interpretations},
  author={Goceri, Evgin},
  journal={Artificial intelligence review},
  volume={56},
  number={11},
  pages={12561--12605},
  year={2023},
  publisher={Springer}
}

@article{blahova2025neural,
  title={Neural Network-Based Mammography Analysis: Augmentation Techniques for Enhanced Cancer Diagnosis—A Review},
  author={Blahov{\'a}, Linda and Kostoln{\`y}, Jozef and Cimr{\'a}k, Ivan},
  journal={Bioengineering},
  volume={12},
  number={3},
  pages={232},
  year={2025}
}

@article{oza2022image,
  title={Image augmentation techniques for mammogram analysis},
  author={Oza, Parita and Sharma, Paawan and Patel, Samir and Adedoyin, Festus and Bruno, Alessandro},
  journal={Journal of Imaging},
  volume={8},
  number={5},
  pages={141},
  year={2022},
  publisher={MDPI}
}

@article{jimenez2024gan,
  title={Gan-based data augmentation to improve breast ultrasound and mammography mass classification},
  author={Jim{\'e}nez-Gaona, Yuliana and Carri{\'o}n-Figueroa, Diana and Lakshminarayanan, Vasudevan and Rodr{\'\i}guez-{\'A}lvarez, Mar{\'\i}a Jos{\'e}},
  journal={Biomedical Signal Processing and Control},
  volume={94},
  pages={106255},
  year={2024},
  publisher={Elsevier}
}

@article{mabotuwana2019automated,
  title={Automated tracking of follow-up imaging recommendations},
  author={Mabotuwana, Thusitha and Hall, Christopher S and Hombal, Vadiraj and Pai, Prashanth and Raghavan, Usha Nandini and Regis, Shawn and McKee, Brady and Dalal, Sandeep and Wald, Christoph and Gunn, Martin L},
  journal={American Journal of Roentgenology},
  volume={212},
  number={6},
  pages={1287--1294},
  year={2019},
  publisher={American Roentgen Ray Society}
}

@inproceedings{jahangir2025mammo,
  title={Mammo-Find: An LLM-Based Multi-channel Tool for Recommending Public Mammogram Datasets},
  author={Jahangir, Raiyan and Filkov, Vladimir},
  booktitle={International Conference on Software Engineering of Emerging Technology},
  pages={446--463},
  year={2025},
  organization={Springer}
}

@inproceedings{jahangir2023proposing,
  title={Proposing novel recurrent neural network architectures for infant cry detection in domestic context},
  author={Jahangir, Raiyan and Mohim, Nasif Shahriar and Khan, Nafiz Imtiaz and Akhtaruzzaman, Md and Islam, Muhammad Nazrul},
  booktitle={2023 IEEE 11th Region 10 Humanitarian Technology Conference (R10-HTC)},
  pages={7--12},
  year={2023},
  organization={IEEE}
}

@article{majib2021vgg,
  title={Vgg-scnet: A vgg net-based deep learning framework for brain tumor detection on mri images},
  author={Majib, Mohammad Shahjahan and Rahman, Md Mahbubur and Sazzad, TM Shahriar and Khan, Nafiz Imtiaz and Dey, Samrat Kumar},
  journal={IEEE Access},
  volume={9},
  pages={116942--116952},
  year={2021},
  publisher={ieee}
}

@inproceedings{khan2020prediction,
  title={Prediction of cesarean childbirth using ensemble machine learning methods},
  author={Khan, Nafiz Imtiaz and Mahmud, Tahasin and Islam, Muhammad Nazrul and Mustafina, Sumaiya Nuha},
  booktitle={Proceedings of the 22nd international conference on information integration and web-based applications \& services},
  pages={331--339},
  year={2020}
}

@inproceedings{rag_3,
  title={Open-Source LLMs for Technical Q\&A: Lessons from StackExchange},
  author={Babar, Zeerak and Khan, Nafiz Imtiaz and Hassnain, Muhammad and Filkov, Vladimir},
  booktitle={International Conference on Software Engineering of Emerging Technology},
  pages={615--626},
  year={2025},
  organization={Springer}
}

@inproceedings{rag_1,
  title={Evidencebot: a privacy-preserving, customizable rag-based tool for enhancing large language model interactions},
  author={Khan, Nafiz Imtiaz and Filkov, Vladimir},
  booktitle={Proceedings of the 33rd ACM International Conference on the Foundations of Software Engineering},
  pages={1188--1192},
  year={2025}
}

@article{rag_2,
  title={Leveraging Language Models to Discover Evidence-Based Actions for OSS Sustainability},
  author={Khan, Nafiz Imtiaz and Filkov, Vladimir},
  journal={arXiv preprint arXiv:2602.11746},
  year={2026}
}

@article{khan2026intelligent,
  title={Intelligent Documentation in Medical Education: Can AI Replace Manual Case Logging?},
  author={Khan, Nafiz Imtiaz and Cleland, Kylie and Filkov, Vladimir and Goldman, Roger Eric},
  journal={arXiv preprint arXiv:2601.12648},
  year={2026}
}

@inproceedings{ramadan2023enhancing,
  title={Enhancing mango leaf disease classification: vit, bit, and cnn-based models evaluated on cyclegan-augmented data},
  author={Ramadan, Syed Taha Yeasin and Sakib, Tanjim and Rahat, Md Ahsan and Mosharrof, Shakil and Rakin, Fatin Ishrak and Jahangir, Raiyan},
  booktitle={2023 26th international conference on computer and information technology (ICCIT)},
  pages={1--6},
  year={2023},
  organization={IEEE}
}

\newpage

\onecolumn
\section*{Appendix}

\begin{lstlisting}[]
Zero Shot Prompt:

You are a board-certified breast radiologist with lots of experience in interpreting screening and diagnostic mammograms. You are meticulous, up to date with the latest BI-RADS guidelines, and always provide clear, concise, and clinically actionable reports.

I am providing you with a mammogram image. The image has all 4 breast views of a patient shown together. The upper two views are the craniocaudal (CC) views of each breast, right and left, and the lower two views are the mediolateral oblique (MLO) views of each breast, right and left. Your task is to analyze the image and provide a structured report in JSON format.
                
For analyzing the image, at first glance, you should identify the breast density using the ACR classification, which includes:
    - ACR A: Almost entirely fatty
    - ACR B: Scattered fibroglandular densities
    - ACR C: Heterogeneously dense
    - ACR D: Extremely dense
                
Then, you should write the abnormalities you notice from the images in 1 sentence. Mention in which view the findings are present, and say "Healthy Breast. No Findings" for normal breasts. Findings include Mass, Suspicious Calcification, Architectural Distortion, Asymmetry, Focal Asymmetry, Global Asymmetry, Suspicious Lymph Nodes, Nipple Retraction, Skin Retraction, Skin Thickening. There may be multiple findings in a single image.
                
Assign a BI-RADS category based on the findings:
    - BI-RADS 1: Negative (no abnormalities)
    - BI-RADS 2: Benign (no suspicion of cancer)
    - BI-RADS 3: Probably benign (short-term follow-up recommended)
    - BI-RADS 4: Suspicious abnormality (biopsy needed)
    - BI-RADS 5: Highly suggestive of malignancy (high probability of cancer)
                
Finally, indicate whether the image is healthy, benign, or suspicious.
                
Here is the JSON format you should follow for your response:
{
    "breast_density": "<ACR A|B|C|D> followed by a brief description of the density",
    "findings": "<Summary of any abnormalities as described above in one sentence>",
    "BI-RADS": "<1|2|3|4|5> followed by a brief description of the BI-RADS category>",
    "suspicion": "<healthy|benign|suspicious>"
}

\end{lstlisting}

\begin{lstlisting}[]
Few-shot Prompt:

<Prompt from Zero-Shot> +
Here are some examples of doctor-annotated reports to guide you:
Example 1:
{
  "image_id": "image_file_path_1",
  "breast_density": "Density C - Heterogeneously Dense. More of the breast is made of dense glandular and fibrous tissue. This can make it hard to see small masses in or around the dense tissue, which also appear as white areas.",
  "BI-RADS": "BI-RADS 1 - Negative. Healthy Breast.",
  "findings": "Healthy Breast. No Findings",
  "suspicion": "healthy"
}

Example 2:
{
  "image_id": "image_file_path_2",
  "breast_density": "Density C - Heterogeneously Dense. More of the breast is made of dense glandular and fibrous tissue. This can make it hard to see small masses in or around the dense tissue, which also appear as white areas.",
  "BI-RADS": "BI-RADS 2 - Benign (non-cancerous) finding",
  "findings": "Healthy Breast. No Findings",
  "suspicion": "healthy"
}
....

\end{lstlisting}

\begin{lstlisting}[]
Chain-Of-Thought Prompt:

<Prompt from Zero-Shot> +

Step 1: Identify breast density
Let's check the breast tissue density. Breast tissue is usually composed of fatty tissue, which appears lighter, and fibro-glandular tissue, which appears darker.

If there is a presence of small white wisps of fibro-glandular strands near the nipple (right and left end in the CC view, lower right and lower left end in the MLO view) against a background of fat, that only covers a few places of the breast. The density is "DENSITY A \- Almost all fatty tissue." Such images are easier to use for diagnosing abnormalities.

If a few scattered pockets of white fibro-glandular tissue start near the nipple and go more outwards, but most of the breast remains of lighter white fat. Those isolated denser areas occupy roughly one-quarter to one-half of the breast, then the density is "DENSITY B - Scattered areas of dense glandular and fibrous tissue". Such images are harder to diagnose any abnormalities.

If there are large swaths of the breast composed of dense, white tissue that start near the nipple, are heterogeneously distributed and patchy, and cover over half the breast, interspersed with lighter fatty areas. Then the density is "DENSITY C \- Heterogeneously Dense. More of the breast is made of dense glandular and fibrous tissue." So, some small findings could be masked and difficult to diagnose.

If the entire breast appears almost uniformly dense white, with very little fat visible,  then the density is "DENSITY D - Extremely Dense. Hard to see masses or other findings that may appear as white areas on the mammogram."  This "extremely dense" pattern of fibroglandular tissue makes it most challenging to detect subtle lesions and diagnose.


Step 2: Now it's time to analyze the breast and identify any abnormalities.

Step 2a: Finding masses or lesions
Let's analyze all views one by one, starting from the right CC view, right MLO view, left CC view, and left MLO view. I scan from top to bottom in a breast view. A mass or lesion should appear as a discrete area of dense white opacity, shaped either round, oval, or irregular, that is visible from both the CC and MLO views of the breast. If it is not visible in one, it may be an overlap or summation artifact. If the mass is well-defined, round, or oval, it is usually benign and should be assigned a BI-RADS 3. If the shape is indistinct and obscured, it raises suspicion of malignancy and should be assigned a BI-RADS 4. If the shape is spiculated or has radiating lines, it is highly suspicious for malignancy and should be assigned a BI-RADS 5. If nothing is visible, no mass is there. Otherwise, I should specify that mass is found in which view of which breast.

Step 2b: Finding calcification
Again, let's analyze each view of the breast from top to bottom. Calcifications appear as tiny white spots on mammograms. Locate all areas of increased density (tiny white specks), punctate, micronodular bright spots on the CC and MLO views. Next, count how many specks are clustered within approximately 1 cubic cm of tissue. Then the morphology of those specks is to be observed. That is, if the specks are round and uniform, or pleomorphic, or have fine-linear branching. Then their distribution has to be ascertained. Are they scattered or grouped in clusters or segmented along ductal anatomy? Then compare with the contralateral breast to assess the asymmetry.

Finally, if the specks are found to be round, "milk-of-calcium," or vascular, they are benign. They are to be assigned BI-RADS 3. If they are pleomorphic or clustered, then assign BI-RADS 4. If the specks are fine-linear or branching, then assign BI-RADS 5.

Step 2c: Finding asymmetry
For determining asymmetry, first line up the CC views of the right and left breasts (and then the MLOs). The target will be to look for areas of density or architectural patterns that appear on one side but not the other. If such a pattern is found in one breast, note its exact position (quadrant, depth) and check the same spot on the other breast and the other views. 

If the asymmetry is visible upon only one projection (either CC or MLO), then it is an asymmetry. An additional image might be taken from a different angle to ascertain. If the asymmetry is persistent but benign, such as a fat lobule, assign BI-RADS 3. However, if suspicious margins/architecture are observed, assign BI-RADS 4.

If the asymmetry is observed in two projections and lacks convex borders or conspicuity of a true mass, and the area covered by the asymmetry is less than or equal to one quadrant in size, it could be a focal asymmetry. It could be assigned BI-RADS 3 if no other abnormalities are available. Based on the presence of mass and suspicious calcification, and other abnormalities, BI-RADS 4 or BI-RADS 5 could be assigned.

If the asymmetry is observed in two projections and spans more than one quadrant, it is a global asymmetry. It could be assigned BI-RADS 3 if no other abnormalities are available. Based on the presence of mass and suspicious calcification, and other abnormalities, BI-RADS 4 or BI-RADS 5 could be assigned.

Step 2d: Finding architectural distortion
Architectural distortion is one of the more subtle but highly important mammographic signs of malignancy. It refers to the disruption of the normal fibro-glandular framework without a discrete mass. To detect them, first identify any suspicious patterns like spiculations or radiating lines with no central mass or focal retraction or pulling of Cooper's ligaments toward a point, or a distortion of parenchymal lines, that is, tissue planes appear bent, kinked, or tethered. If the distortion is stable, assign BI-RADS 3. If the distortion is mild, with sparse striations or dense, thick spicules, assign BI-RADS 4. Else if it has a classic starburst pattern, assign BI-RADS 5.

Step 2e: Check for suspicious lymph nodes
The axillary lymph nodes lie in the upper outer quadrant of each breast. It is normally visible in the MLO view. In the given image, it is in the upper central of the MLO view. If nodes are visible and reniform (bean-shaped) with a long axis parallel to the skin, then it is benign. Assign BI-RADS 2. If mild, diffuse cortical thickening without other worrisome features is observed, it is likely benign. Assign BI-RADS 3. If it is round-shaped with a focal cortical bulge or loss of hilum, it is highly suspicious of malignancy. Assign BI-RADS 4. If the cortex is markedly thickened with spiculated margins or clustered microcalcifications, biopsy is urgent. Assign BI-RADS 5.

Step 2f: Finding other observations
First, check the nipple retraction. It is an inward pulling or inversion of the nipple. At first, in the CC view, look at the nipple border. Normally, it projects forward as a small convex contour.
Retraction shows as an inward indentation or loss of the convex silhouette. Then check the MLO view and confirm that the nipple tip lies posterior (toward the chest wall) relative to the skin line rather than anterior.

Then check skin retraction. It is a focal pulling in of the skin surface, often from an underlying desmoplastic (fibrotic) reaction. Search for a subtle focal area where skin thickness suddenly narrows or a dimple forms, often overlying a spiculated mass. Also, look for converging parenchymal lines (ligamentous strands) that course from the lesion toward the skin. Then, verify if it is on two projections. 

Then check skin thickening. It is a diffuse or focal increase in the thickness of the subcutaneous fat layer, which can be due to edema, an inflammatory cancer (e.g., Paget's disease, inflammatory carcinoma), or prior surgery/radiation. At first, identify the skin line on both CC and MLO views. Normally, the skin appears as a thin radiopaque line ~1.5 to 2mm thick. If it is a Diffuse thickening >2.5mm (some texts use >3mm) across multiple quadrants, it is abnormal. If it is a focal thickening >3mm in a localized area, it warrants further work-up. Bilateral and symmetric thickening often reflects systemic edema (e.g., heart failure). Unilateral or focal suggests underlying inflammatory malignancy or local process. Trabecular thickening (edema) in the subcutaneous fat or Cooper's ligaments. Usually, such features do not indicate malignancy and may be considered BI-RADS 1 or BI-RADS 2 in the presence of multiple observations. If observed with other abnormalities, they may be assigned a higher BI-RADS score.

Step 2g: If nothing is found,
If nothing is found, write "Healthy Breast. No Findings" and assign BI-RADS 1.

Step 2h: Write the findings 
Write which abnormalities are found in which view of which breast. If there are multiple abnormalities, write them all.

Step 3: Assign a BI-RADS Score
Based on the above explanation of findings, assign a BI-RADS score. 

                - BI-RADS 0: Incomplete (needs additional imaging)
                - BI-RADS 1: Negative (no abnormalities)
                - BI-RADS 2: Benign (no suspicion of cancer)
                - BI-RADS 3: Probably benign (short-term follow-up recommended)
                - BI-RADS 4: Suspicious abnormality (biopsy needed)
                - BI-RADS 5: Highly suggestive of malignancy (high probability of cancer)
                - BI-RADS 6: Known malignancy (biopsy-proven cancer) 


Step 4: Finally, indicate whether the case is healthy, benign, or suspicious (radiologic suspicion, not biopsy-confirmed cancer).

Step 5: Here is the JSON format you should follow for your response:
{
    "breast_density": "<ACR A|B|C|D> followed by a brief description of the density",
    "findings": "<Summary of any abnormalities as described above in one sentence>",
    "BI-RADS": "<1|2|3|4|5> followed by a brief description of the BI-RADS category>",
    "suspicion": "<healthy|benign|suspicious>"
}
\end{lstlisting}

% Please add the following required packages to your document preamble:
% \usepackage{multirow}
% \usepackage{graphicx}
% \usepackage{booktabs}

\begin{table*}[!ht]
\centering
\caption{Classification performance of different models on VinDr-Mammo dataset on zero-shot}
\label{zeroshot-vindr-class}
\resizebox{\columnwidth}{!}{%
\begin{tabular}{llccc}
\toprule
\multirow{2}{*}{\textbf{Task}} &
  \multirow{2}{*}{\textbf{Evaluation Parameter}} &
  \multicolumn{3}{c}{\textbf{Models}} \\
\cmidrule(lr){3-5}
 & & \textbf{MedGemma} & \textbf{Qwen2.5VL} & \textbf{LLAVA-Med} \\
\midrule

\multirow{5}{*}{BI-RADS}
 & Accuracy    & 0.2218 & 0.1957 & 0.2273 \\
 & Precision   & 0.2722 & 0.0855 & 0.0517 \\
 & Recall      & 0.2218 & 0.1957 & 0.2273 \\
 & F1-score    & 0.1237 & 0.0913 & 0.0842 \\
 & Specificity & 0.8006 & 0.8008 & 0.8    \\
\midrule

\multirow{5}{*}{Breast Density}
 & Accuracy    & 0.7173 & 0.4336 & 0.1041 \\
 & Precision   & 0.6507 & 0.6223 & 0.7925 \\
 & Recall      & 0.7173 & 0.4336 & 0.1041 \\
 & F1-score    & 0.6819 & 0.5105 & 0.0211 \\
 & Specificity & 0.7754 & 0.7587 & 0.7503 \\
\midrule

\multirow{5}{*}{Calcification}
 & Accuracy    & 0.4009 & 0.5282 & 0.8368 \\
 & Precision   & 0.7461 & 0.7528 & 0.7003 \\
 & Recall      & 0.4009 & 0.5282 & 0.8368 \\
 & F1-score    & 0.4535 & 0.5887 & 0.7625 \\
 & Specificity & 0.5254 & 0.5477 & 0.5    \\
\midrule

\multirow{5}{*}{Mass}
 & Accuracy    & 0.4945 & 0.6186 & 0.6191 \\
 & Precision   & 0.5272 & 0.3832 & 0.3833 \\
 & Recall      & 0.4945 & 0.6186 & 0.6191 \\
 & F1-score    & 0.5016 & 0.4732 & 0.4734 \\
 & Specificity & 0.4988 & 0.4996 & 0.5    \\
\midrule

\multirow{5}{*}{Asymmetry}
 & Accuracy    & 0.8545 & 0.5095 & 0.8915 \\
 & Precision   & 0.7302 & 0.7587 & 0      \\
 & Recall      & 0.8545 & 0.5095 & 0      \\
 & F1-score    & 0.7875 & 0.5802 & 0      \\
 & Specificity & 0.5    & 0.5147 & 1      \\
\midrule

\multirow{5}{*}{Suspicion}
 & Accuracy    & 0.3464 & 0.6396 & 0.6818 \\
 & Precision   & 0.6325 & 0      & 0.4649 \\
 & Recall      & 0.3464 & 0      & 0.6818 \\
 & F1-score    & 0.2277 & 0      & 0.5528 \\
 & Specificity & 0.5111 & 1      & 0.5    \\
\bottomrule
\end{tabular}%
}
\end{table*}
% Please add the following required packages to your document preamble:
% \usepackage{multirow}
% \usepackage{graphicx}
% \usepackage{booktabs}

\begin{table*}[!ht]
\centering
\caption{Classification performance of different models on the VinDr-Mammo dataset on few-shot}
\label{fewshot-vindr-class}
\resizebox{\columnwidth}{!}{%
\begin{tabular}{llccc}
\toprule
\multirow{2}{*}{\textbf{Task}} &
  \multirow{2}{*}{\textbf{Evaluation Parameter}} &
  \multicolumn{3}{c}{\textbf{Models}} \\
\cmidrule(lr){3-5}
 & & \textbf{MedGemma} & \textbf{Qwen2.5VL} & \textbf{LLAVA-Med} \\
\midrule

\multirow{5}{*}{BI-RADS}
 & Accuracy    & 0.1019 & 0.2582 & 0.2909 \\
 & Precision   & 0.1515 & 0.4121 & 0.3493 \\
 & Recall      & 0.1019 & 0.2582 & 0.2909 \\
 & F1-score    & 0.0376 & 0.1876 & 0.2217 \\
 & Specificity & 0.8004 & 0.808  & 0.8165 \\
\midrule

\multirow{5}{*}{Breast Density}
 & Accuracy    & 0.5921 & 0.7832 & 0.1141 \\
 & Precision   & 0.6818 & 0.7153 & 0.6446 \\
 & Recall      & 0.5921 & 0.7832 & 0.1141 \\
 & F1-score    & 0.6061 & 0.6893 & 0.046  \\
 & Specificity & 0.7649 & 0.7516 & 0.7506 \\
\midrule

\multirow{5}{*}{Calcification}
 & Accuracy    & 0.2055 & 0.8391 & 0.8377 \\
 & Precision   & 0.7725 & 0.865  & 0.8235 \\
 & Recall      & 0.2055 & 0.8391 & 0.8377 \\
 & F1-score    & 0.1368 & 0.7679 & 0.7655 \\
 & Specificity & 0.5107 & 0.507  & 0.5039 \\
\midrule

\multirow{5}{*}{Mass}
 & Accuracy    & 0.4127 & 0.6205 & 0.6195 \\
 & Precision   & 0.6556 & 0.7647 & 0.7644 \\
 & Recall      & 0.4127 & 0.6205 & 0.6195 \\
 & F1-score    & 0.2885 & 0.4766 & 0.4745 \\
 & Specificity & 0.5211 & 0.5018 & 0.5006 \\
\midrule

\multirow{5}{*}{Asymmetry}
 & Accuracy    & 0.8259 & 0.7159 & 0.7514 \\
 & Precision   & 0.7359 & 0.7608 & 0.761  \\
 & Recall      & 0.8259 & 0.7159 & 0.7514 \\
 & F1-score    & 0.7759 & 0.7362 & 0.7561 \\
 & Specificity & 0.4884 & 0.5213 & 0.52   \\
\midrule

\multirow{5}{*}{Suspicion}
 & Accuracy    & 0.3291 & 0.6841 & 0.5914 \\
 & Precision   & 0.7579 & 0.7841 & 0.5881 \\
 & Recall      & 0.3291 & 0.6841 & 0.5914 \\
 & F1-score    & 0.1771 & 0.5581 & 0.5897 \\
 & Specificity & 0.5076 & 0.5036 & 0.5251 \\
\bottomrule
\end{tabular}%
}
\end{table*}
% Please add the following required packages to your document preamble:
% \usepackage{multirow}
% \usepackage{graphicx}
% \usepackage{booktabs}

\begin{table*}[!ht]
\centering
\caption{Classification performance of different models on VinDr-Mammo dataset on chain-of-thought}
\label{cot-vindr-class}
\resizebox{\columnwidth}{!}{%
\begin{tabular}{llccc}
\toprule
\multirow{2}{*}{\textbf{Task}} &
  \multirow{2}{*}{\textbf{Evaluation Parameter}} &
  \multicolumn{3}{c}{\textbf{Models}} \\
\cmidrule(lr){3-5}
 & & \textbf{MedGemma} & \textbf{Qwen2.5VL} & \textbf{LLAVA-Med} \\
\midrule

\multirow{5}{*}{BI-RADS}
 & Accuracy    & 0.2493 & 0.2291 & 0.2488 \\
 & Precision   & 0.1846 & 0.0526 & 0.1898 \\
 & Recall      & 0.2493 & 0.2291 & 0.2488 \\
 & F1-score    & 0.1412 & 0.0855 & 0.2059 \\
 & Specificity & 0.8059 & 0.8    & 0.8046 \\
\midrule

\multirow{5}{*}{Breast Density}
 & Accuracy    & 0.742  & 0.0168 & 0.0386 \\
 & Precision   & 0.6631 & 0.7923 & 0.0086 \\
 & Recall      & 0.742  & 0.0168 & 0.0386 \\
 & F1-score    & 0.69   & 0.0155 & 0.014  \\
 & Specificity & 0.7667 & 0.7506 & 0.7475 \\
\midrule

\multirow{5}{*}{Calcification}
 & Accuracy    & 0.6641 & 0.1695 & 0.525  \\
 & Precision   & 0.7466 & 0.689  & 0.7164 \\
 & Recall      & 0.6641 & 0.1695 & 0.525  \\
 & F1-score    & 0.6973 & 0.0626 & 0.587  \\
 & Specificity & 0.5414 & 0.4982 & 0.4796 \\
\midrule

\multirow{5}{*}{Mass}
 & Accuracy    & 0.4914 & 0.6191 & 0.6218 \\
 & Precision   & 0.5305 & 0.3833 & 0.6273 \\
 & Recall      & 0.4914 & 0.6191 & 0.6218 \\
 & F1-score    & 0.4976 & 0.4734 & 0.486  \\
 & Specificity & 0.5022 & 0.5    & 0.5054 \\
\midrule

\multirow{5}{*}{Asymmetry}
 & Accuracy    & 0.8545 & 0.1505 & 0.5723 \\
 & Precision   & 0.7302 & 0.6362 & 0.7611 \\
 & Recall      & 0.8545 & 0.1505 & 0.5723 \\
 & F1-score    & 0.7875 & 0.0526 & 0.635  \\
 & Specificity & 0.5    & 0.4938 & 0.5215 \\
\midrule

\multirow{5}{*}{Suspicion}
 & Accuracy    & 0.3982 & 0.6832 & 0.3279 \\
 & Precision   & 0.6464 & 0.4668 & 0.4435 \\
 & Recall      & 0.3982 & 0.6832 & 0.3279 \\
 & F1-score    & 0.3369 & 0.5546 & 0.1629 \\
 & Specificity & 0.5347 & 0.5    & 0.4995 \\
\bottomrule
\end{tabular}%
}
\end{table*}

% Please add the following required packages to your document preamble:
% \usepackage{multirow}
% \usepackage{graphicx}
% \usepackage{booktabs}

\begin{table*}[!ht]
\centering
\caption{Performance of different models on DMID dataset with zero-shot prompting}
\label{zeroshot-dmid}
\resizebox{\columnwidth}{!}{%
\begin{tabular}{llccc}
\toprule
\textbf{Task} & \textbf{Evaluation Parameters} & \textbf{MedGemma} & \textbf{LLAVA-Med} & \textbf{Qwen2.5VL} \\
\midrule

\multirow{7}{*}{BI-RADS}
 & Accuracy    & 0.2878 & 0.2    & 0.2918 \\
 & Precision   & 0.2861 & 0.0824 & 0.6562 \\
 & Recall      & 0.2878 & 0.2    & 0.2918 \\
 & F1-score    & 0.2602 & 0.1167 & 0.247  \\
 & Specificity & 0.8371 & 0.8    & 0.8268 \\
 & BERTScore   & 0.8662 & 0.8322 & 0.8557 \\
 & ROUGE-L     & 0.1766 & 0.1373 & 0.0492 \\
\midrule

\multirow{7}{*}{Breast Density}
 & Accuracy    & 0.2649 & 0.25   & 0.268  \\
 & Precision   & 0.3077 & 0.0995 & 0.1424 \\
 & Recall      & 0.2649 & 0.25   & 0.268  \\
 & F1-score    & 0.2476 & 0.1424 & 0.1782 \\
 & Specificity & 0.7554 & 0.75   & 0.7585 \\
 & BERTScore   & 0.8408 & 0.8408 & 0.8217 \\
 & ROUGE-L     & 0.1984 & 0.2228 & 0.216  \\
\midrule

\multirow{2}{*}{Findings}
 & BERTScore   & 0.843  & 0.8293 & 0.8252 \\
 & ROUGE-L     & 0.1437 & 0.0861 & 0.1298 \\
\bottomrule
\end{tabular}%
}
\end{table*}
% Please add the following required packages to your document preamble:
% \usepackage{multirow}
% \usepackage{graphicx}
% \usepackage{booktabs}

\begin{table*}[!ht]
\centering
\caption{Performance of different models on DMID dataset with few-shot prompting}
\label{fewshot-dmid}
\resizebox{\columnwidth}{!}{%
\begin{tabular}{llccc}
\toprule
\textbf{Task} & \textbf{Evaluation Parameters} & \textbf{MedGemma} & \textbf{LLAVA-Med} & \textbf{Qwen2.5VL} \\
\midrule

\multirow{7}{*}{BI-RADS}
 & Accuracy    & 0.3531 & 0.3998 & 0.201  \\
 & Precision   & 0.3717 & 0.8139 & 0.2483 \\
 & Recall      & 0.3531 & 0.3998 & 0.201  \\
 & F1-score    & 0.3392 & 0.3419 & 0.0797 \\
 & Specificity & 0.8542 & 0.8602 & 0.8005 \\
 & BERTScore   & 0.9983 & 0.9508 & 0.9999 \\
 & ROUGE-L     & 0.4906 & 0.0392 & 0.2431 \\
\midrule

\multirow{7}{*}{Breast Density}
 & Accuracy    & 0.3739 & 0.2226 & 0.2512 \\
 & Precision   & 0.3685 & 0.3664 & 0.3423 \\
 & Recall      & 0.3739 & 0.2226 & 0.2512 \\
 & F1-score    & 0.3605 & 0.155  & 0.1373 \\
 & Specificity & 0.7877 & 0.7429 & 0.7508 \\
 & BERTScore   & 0.8991 & 0.8093 & 0.8862 \\
 & ROUGE-L     & 0.4939 & 0.1726 & 0.3036 \\
\midrule

\multirow{2}{*}{Findings}
 & BERTScore   & 0.8615 & 0.8141 & 0.8492 \\
 & ROUGE-L     & 0.2706 & 0.1076 & 0.2303 \\
\bottomrule
\end{tabular}%
}
\end{table*}
% Please add the following required packages to your document preamble:
% \usepackage{multirow}
% \usepackage{graphicx}
% \usepackage{booktabs}

\begin{table*}[!ht]
\centering
\caption{Performance of different models on DMID dataset with Chain-of-Thought prompting}
\label{cot-dmid}
\resizebox{\columnwidth}{!}{%
\begin{tabular}{llccc}
\toprule
\textbf{Task} & \textbf{Evaluation Parameters} & \textbf{MedGemma} & \textbf{LLAVA-Med} & \textbf{Qwen2.5VL} \\
\midrule

\multirow{7}{*}{BI-RADS}
 & Accuracy    & 0.2277 & 0.183  & 0.57   \\
 & Precision   & 0.3108 & 0.336 & 0.65   \\
 & Recall      & 0.2277 & 0.183  & 0.57   \\
 & F1-score    & 0.1462 & 0.1288 & 0.59   \\
 & Specificity & 0.81   & 0.8358 & 0.77   \\
 & BERTScore   & 0.8657 & 0.9939 & 0.8448 \\
 & ROUGE-L     & 0.0373 & 0.3982 & 0.0382 \\
\midrule

\multirow{7}{*}{Breast Density}
 & Accuracy    & 0.2823 & 0.2494 & 0.35   \\
 & Precision   & 0.3361 & 0.1168 & 0.65   \\
 & Recall      & 0.2823 & 0.2494 & 0.35   \\
 & F1-score    & 0.2711 & 0.1082 & 0.3    \\
 & Specificity & 0.7674 & 0.7477 & 0.7316 \\
 & BERTScore   & 0.8508 & 0.7696 & 0.72   \\
 & ROUGE-L     & 0.1582 & 0.0632 & 0.2    \\
\midrule

\multirow{2}{*}{Findings}
 & BERTScore   & 0.8478 & 0.7702 & 0.55 \\
 & ROUGE-L     & 0.1388 & 0.0405 & 0.12 \\
\bottomrule
\end{tabular}%
}
\end{table*}

% % References
% %==============================================
% \bibliographystyle{vancouver}
% % {\footnotesize \bibliography{citation}}
% \bibliography{reference}

% \onecolumn
% \include{appendix}

\end{document}